\documentclass{article}

\usepackage[main, final]{neurips_2026}
\workshoptitle{Dynamic Alignment on Human-AI Coupled System (BiAlign Workshop)}

\usepackage{wrapfig}

\usepackage{array}
\usepackage{makecell}
\newcommand{\headcell}[1]{\makebox[4.8em][c]{\textbf{#1}}}
\newcommand{\eqcell}[2]{\cellcolor{#1}{\makebox[7em][c]{#2}}}
\usepackage[table]{xcolor}

\usepackage[utf8]{inputenc} 
\usepackage[T1]{fontenc}    
\usepackage{hyperref}       
\usepackage{url}            
\usepackage{booktabs}       
\usepackage{amsfonts}       
\usepackage{amsmath}
\usepackage{nicefrac}       
\usepackage{microtype}      
\usepackage[table]{xcolor}
\usepackage{xcolor}         
\usepackage{diagbox}
\usepackage{tikz}
\usepackage{caption}
\usepackage{multirow}
\usepackage{soul}
\usepackage{booktabs,graphicx,makecell}
\providecommand{\headcell}[1]{\makecell{\textbf{#1}}}

\title{Probing Human–Machine Perceptual Alignment with Semantically Ambiguous Images}

\author{
Yuqi Hu$^1$\qquad
Vasha DuTell$^{\ast 1,2}$\qquad
Ahna R. Girshick$^{\ast1}$\qquad
Jennifer E. Corbett$^{\ast2}$\qquad
\\[0.5em]
$^1$University of California, Berkeley \ \ \ 
$^2$Massachusetts Institute of Technology \ \ \
\\
\normalsize
\{yuqihu,~ahna,~vasha\}@berkeley.edu \ \ \ 
\{vasha,~j8675309\}@mit.edu 
\\[0.5em]
* Equal Contribution
}

\begin{document}

\maketitle

\begin{abstract}

The classic duck–rabbit illusion reveals a fundamental property of perception: when visual evidence is ambiguous, a visual system must resolve competing interpretations. But do humans and machine classifiers resolve the same ambiguity in the same way? To explore this, we use semantically ambiguous images as probes of human–machine perceptual alignment. Our psychophysically-informed framework interpolates between concepts in vision-language embedding space to generate continuous spectra of ambiguous images, enabling direct comparison of human and machine perceptual decision boundaries and sensitivities. We find systematic differences in these boundaries: machine classifiers are shifted relative to both human observers and the midpoint of the synthesis continuum. Guidance scale affects both bias and sensitivity in humans and the machine classifiers tested. Controlled semantic ambiguity in generated images, combined with a psychophysical framework, can thus enable quantitative evaluation of human–machine perceptual alignment.

\end{abstract}    
\section{Introduction and Previous Work}
\label{sec:intro}

\begin{wrapfigure}[13]{r}{0.34\textwidth}
    \vspace{-1.5\baselineskip}
    \centering
    \includegraphics[width=0.9\linewidth]{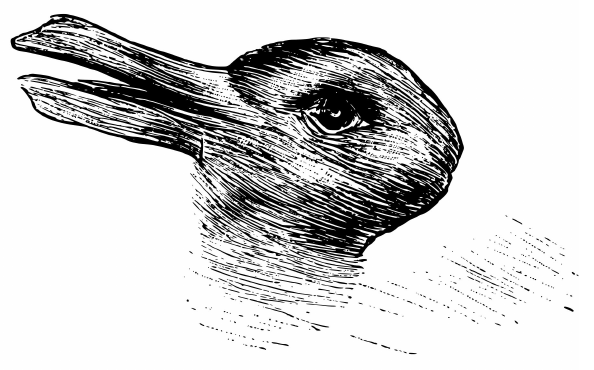}
    \caption{The classic duck–rabbit illusion~\citep{jastrow1899duck} is a well-known example of perceptual ambiguity, where the same image can be seen as either a duck or a rabbit.}
    \label{fig:duck_rabbit_classic}
    \vspace{0.5\baselineskip}
\end{wrapfigure}

The classic duck–rabbit illusion ~\citep{jastrow1899duck} (Fig.~\ref{fig:duck_rabbit_classic}) illustrates a striking property of perception: a single image can support multiple valid interpretations. How a visual system resolves such ambiguity reveals where it places the boundary between competing perceptual hypotheses. Comparing these boundaries across humans and machine models provides a direct way to probe human–machine perceptual alignment.

Psychophysics~\citep{read2015psychophysics} provides quantitative methods for probing human perception, often using controlled continua of stimuli. Classic bistable stimuli based on low-level visual cues, such as the Necker cube and Rubin’s vase, have long served as probes of perceptual ambiguity~\citep{leopold1999multistable}. In contrast, the duck–rabbit is ambiguous at the level of semantic meaning—a form of indeterminate imagery~\citep{pepperell2006seeing, hertzmann2020visual} that is difficult to parameterize systematically. Text-to-image diffusion models offer a new scalable way to construct continuous stimulus spectra that transition between semantic concepts, enabling quantitative comparison of how humans and machine models resolve the same ambiguity.

\begin{figure}[t]
  \centering
  \includegraphics[width=\linewidth]{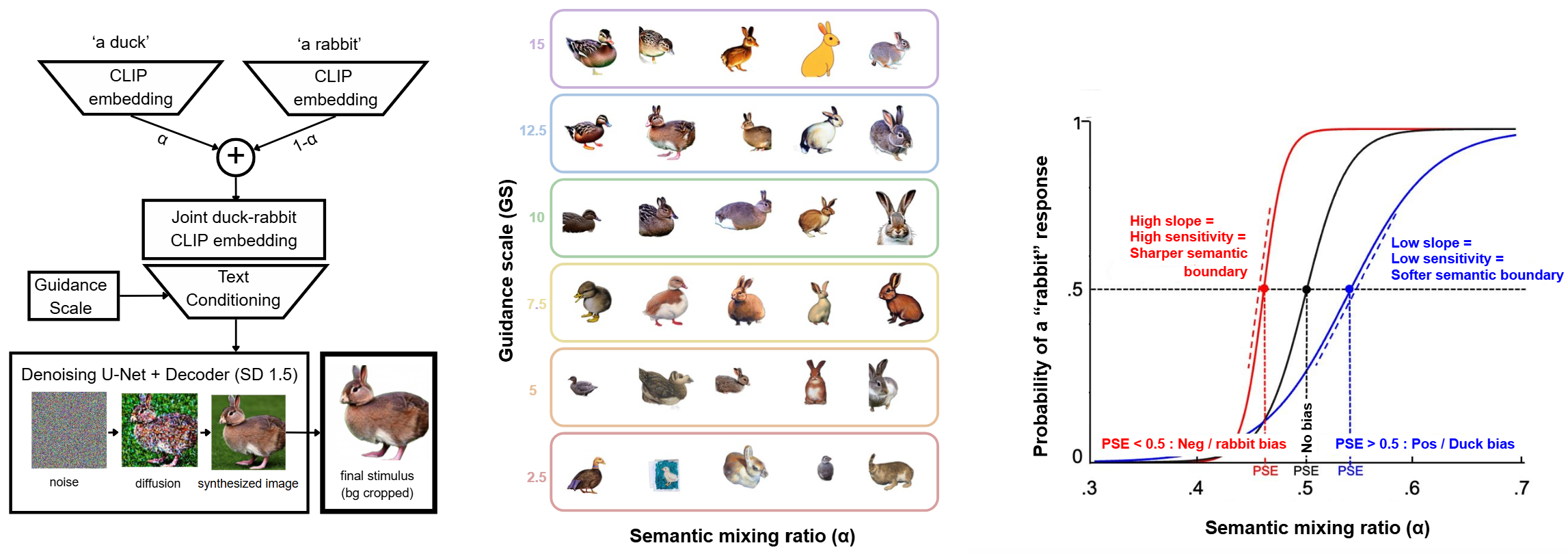}
  \caption{\textbf{Left:} Our process for generating ambiguous stimuli. \textbf{Middle:} Example stimuli as a function of semantic mixing ratio ($\alpha$) and guidance scale (GS). \textbf{Right:} Illustrative psychometric function and effects on bias and sensitivity as a function of $\alpha$. }
  \label{fig:duck_rabbit_process}
\end{figure}

From a machine learning perspective, this comparison matters for understanding the geometry of learned visual representations. Modern machine vision systems, from CLIP~\citep{radford2021learning} to ImageNet classifiers~\citep{russakovsky2015imagenet}, learn representations that support semantic distinctions among visual concepts, yet it remains difficult to characterize where their semantic decision boundaries lie and how they correspond to human perception. Direct comparison is further complicated by differences in human and machine response formats (e.g., 2AFC vs. classifier probabilities).

Here, we introduce a shared psychophysical framework for quantifying human–machine perceptual alignment by comparing perceptual decision boundaries along a controlled ambiguity continuum. We define \emph{perceptual alignment} as the degree of correspondence between human and machine judgments of the same visual evidence, with the CLIP-defined synthesis continuum providing a common reference. By interpolating between text embeddings of a ``duck'' and ``a rabbit'', we generate stimuli spanning a continuous semantic mixing ratio ($\alpha$), while varying classifier-free guidance scale (GS)~\citep{ho2022classifier} to control how strongly image generation is guided by the interpolated text embeddings. We measure judgments from human observers using a two-alternative forced-choice (2AFC) task and from pretrained ImageNet classifiers using aggregated softmax probabilities over relevant semantic categories. Psychometric functions provide common estimates of decision boundary and sensitivity for humans and models at each guidance scale. This framework uses controlled semantic ambiguity to reveal where human and machine perceptual judgments align and systematically diverge.

\section{A Framework for Semantic Ambiguity}
\label{sec:methods}

\subsection{Stimulus Generation Along a Continuum}
The first component of our framework constructs a controlled semantic continuum for comparing human and machine perception. As shown in Fig.~\ref{fig:duck_rabbit_process} (left), we generate stimuli using Stable Diffusion v1.5~\citep{stable-diffusion-v1-5}, based on the latent diffusion formulation~\citep{rombach2022high} with DDPM denoising~\citep{ho2020denoising}, CLIP text conditioning~\citep{radford2021learning}, and classifier-free guidance~\citep{ho2022classifier}. To systematically vary semantic ambiguity, we linearly interpolate between CLIP embeddings using a mixing parameter $\alpha \in [0,1]$. For the duck–rabbit experiment (see Supplementary for replication with elephant–rabbit), the prompts ``a duck'' and ``a rabbit'' are encoded as $\mathrm{emb}_{\text{duck}}$ and $\mathrm{emb}_{\text{rabbit}}$, yielding the linearly interpolated (lerp) conditioning embedding $\mathrm{emb}_{\mathrm{cond}}(\alpha) = (1 - \alpha)\,\mathrm{emb}_{\text{duck}} + \alpha\,\mathrm{emb}_{\text{rabbit}}$. This parameterization provides a common stimulus axis along which human and machine semantic decision boundaries can be measured and compared.

Setting $\alpha$ = 0 produces a duck-dominant image; $\alpha$  = 1 produces a rabbit-dominant image; and in-between values yield hybrid duck-rabbit images. We factorially combined six guidance scales $\{2.5,\,5.0,\,7.5,\,10.0,\,12.5,\,15.0\}$ with five $\alpha$ levels $\{0.3,\,0.4,\,0.5,\,0.6,\,0.7\}$, selected through pilot testing to span the psychometric range.  Ten synthesis seeds per condition produced 300 images with 60 denoising steps each (Fig.~\ref{fig:duck_rabbit_process} (Middle); see Supplementary for the full stimulus set (Sec.~\ref{sec:app_stimuli_generation})).

\subsection{Perceptual Alignment Measurements on a Matched Continuum}
\textbf{Humans.} We measured human responses in a 2AFC experiment (see Supplementary Sec.~\ref{sec:app_psychophysics}) and fit the proportion of ``rabbit'' responses across $\alpha$ with a logistic psychometric function using psignifit 4~\citep{schutt2016painfree}:
$P(\text{rabbit}\mid\alpha) \approx \frac{1}{1 + \exp\!\left[-\beta_1(\alpha - \mathrm{PSE})\right]}.$ 
The point of subjective equality (PSE) defines the semantic decision boundary, where $P(\text{rabbit | $\alpha$ = PSE})=0.5$. We quantify $Bias=\mathrm{PSE}-0.5$, such that zero indicates a boundary at the midpoint of the CLIP-defined continuum, and $Sensitivity \propto \beta_1$, which captures the sharpness of the transition around that boundary (Fig.~\ref{fig:duck_rabbit_process} (Right)). Together, these measures characterize where observers place their semantic boundary and how sharply they distinguish the two concepts.

\textbf{Machine Classifiers.} We evaluated the same 300 images using eight pretrained ImageNet classifiers (see Supplementary Sec.~\ref{sec:supp_classifier_models}). To place machine and human judgments on a matched psychophysical scale, we aggregated classifier softmax probabilities over ImageNet classes corresponding to ``rabbit'' and ``duck'' and normalized them to obtain probability $p_\text{rabbit}$ for each image. Using $p_\text{rabbit}$, we then generated one Bernoulli response per image to create a simulated machine participant, and fit the simulated participant's responses with the same psychometric model used for humans. We then repeated this $n$ times, where $n$ equals the number of human participants. We additionally replicated the framework with an elephant–rabbit continuum in both humans and the same classifiers (Supplementary Sec.~\ref{sec:supp_elephant_rabbit}).

\subsection{Systematic divergence of human and machine perceptual decision boundaries}

\begin{figure*}[t]
  \centering
  \vspace{-1em}
    \includegraphics[width=\linewidth]{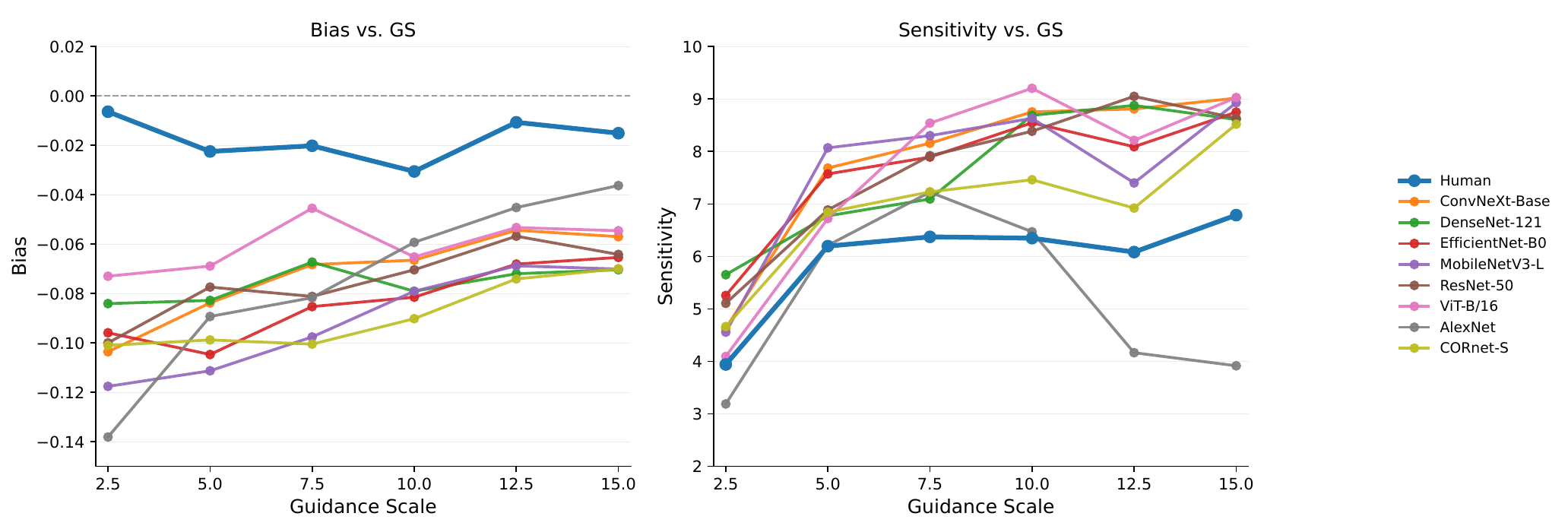}
\caption{\textbf{Human and machine perceptual bias and sensitivity across guidance scales for the duck–rabbit condition.}
\textbf{Left:} Estimated bias; the dashed line indicates the midpoint of the synthesis continuum (zero bias), with positive and negative values corresponding to duck and rabbit bias, respectively.
\textbf{Right:} Higher sensitivity values indicate stronger semantic separation as the mixing parameter $\alpha$ varies.}
  \vspace{-1em}
\label{fig:bias_sensitivity_guidance}
\end{figure*}

\textbf{Framework Validation.} The image synthesis approach produced well-behaved perceptual continua: human and classifier responses transitioned smoothly and largely monotonically from duck to rabbit across $\alpha$, supporting linear interpolation in CLIP embedding space as a method for generating controlled semantic ambiguity. Psychometric functions covered the response range and showed acceptable goodness-of-fit (Fig.~\ref{fig:psychometric_curves_dr} and Table~\ref{tab:deviance_guidance_dr}, Supplementary).

\textbf{Bias.} 
Human and machine classifiers showed somewhat different semantic decision boundaries (Fig.~\ref{fig:bias_sensitivity_guidance}, left; Supplementary Table~\ref{tab:bias_heat} and Fig.~\ref{fig:bias_gs_bar}). In the duck-rabbit experiment, across guidance scales, human bias was closer to the midpoint of the CLIP-defined continuum than machine models. Classifiers showed notably greater rabbit bias across architectures. AlexNet shows the most pronounced effect of guidance scale in reducing rabbit bias. Other classifiers also show this descriptive trend, although further testing is needed to establish this effect.

The elephant–rabbit experiment showed a qualitatively similar human–machine difference in decision boundaries, with humans again less biased than classifiers, although classifier biases were smaller than in the duck–rabbit experiment. AlexNet was again an outlier, this time showing a consistent elephant bias unlike the rest of the models (Table~\ref{tab:bias_er}, Fig.~\ref{fig:bias_sensitivity_guidance_er} and Fig.~\ref{fig:bias_gs_bar_er}, Supplementary). 

\textbf{Sensitivity.} 
In the duck-rabbit experiment, machine classifiers generally showed greater sensitivity than humans, indicating sharper transitions between semantic categories (Fig.~\ref{fig:bias_sensitivity_guidance}, right; Supplementary Table~\ref{tab:slope} and Fig.~\ref{fig:sensitivity_gs_bar}). Humans and most classifiers had the largest increase in sensitivity between GS $=2.5$ and $5$, and then increased only slightly between GS $=5$ and $15$. AlexNet was an exception, with a decrease in sensitivity that put it well below humans at high GS (between $=7.5$ and $15$).

The elephant–rabbit experiment showed a similar overall pattern of greater sensitivity in classifiers than humans (Table~\ref{tab:slope_er}, Fig.~\ref{fig:bias_sensitivity_guidance_er} and Fig.~\ref{fig:sensitivity_gs_bar_er}, Supplementary). As in the duck–rabbit experiment, most classifiers showed increasing sensitivity as GS increased (between GS $=2.5$ and $10$). However, unlike the duck-rabbit experiment, human and all classifiers' sensitivity decreased between GS $=10$ and $15$.
\section{Discussion and Future Directions}
\label{sec:discussion}

A fundamental question in human–machine alignment is whether humans and machines interpret the same visual evidence in the same way, particularly when that evidence is ambiguous. Semantically ambiguous images such as the duck–rabbit illusion provide a natural test case because they require a perceptual system to resolve competing interpretations~\citep{jastrow1899duck, rock1983logic}. Quantitative psychophysical comparison requires systematically varying the evidence between competing interpretations~\citep{read2015psychophysics}. Our framework provides this controlled semantic continuum and applies the same psychophysical analysis to human and machine judgments, enabling quantitative comparison of their perceptual decision boundaries and sensitivities.

Across two conditions, we find that human observers exhibit less semantic bias than classifier models relative to the midpoint of the CLIP-defined synthesis continuum, placing human decision boundaries closer to this reference point. More importantly for human–machine alignment, the results reveal systematic differences between human and classifier decision boundaries when judging the same ambiguous visual evidence. An open question is whether these differences originate primarily from the training data and objectives of the classifiers, or from the statistics of the generated images themselves. Disentangling these factors will be important for understanding the sources of human–machine perceptual misalignment.

We further observe that classifier models tend to exhibit higher sensitivity than humans, potentially reflecting differences in internal noise, decision strategies, or human tendency to fatigue during the experiment~\citep{peterson2019human}. In some cases, both bias and sensitivity vary with guidance scale, suggesting that guidance scale modulates not only image quality but also the structure of ambiguity in generated stimuli. Thus, guidance scale may control the sharpness of semantic transitions, rather than acting only as a fidelity parameter, with effects depending on stimulus class and model architecture. Extending this framework to a wider range of label pairs, such as pairs of ImageNet animals~\citep{russakovsky2015imagenet}, could provide a more comprehensive map of human–machine perceptual alignment.

Interestingly, different classifier architectures show broadly similar behavior despite substantial differences in design. Vision transformers~\citep{dosovitskiy2020image} and convolutional networks~\citep{he2016deep} yield comparable perceptual boundaries, and even architectures motivated by biological vision, such as CORnet~\citep{kubilius2019brain}, do not exhibit substantially more human-like behavior. The oldest of these models, AlexNet~\citep{krizhevsky2012imagenet}, shows occasional deviations, although the source of these differences remains unclear. These results suggest perceptual boundaries may be shaped by factors other than architecture, such as training objectives and training datasets. Future work should extend this analysis to a broader range of models, including vision--language models such as Qwen2-VL, DeepSeek-VL, and PaliGemma~\citep{wang2024qwen2,lu2024deepseek,beyer2024paligemma}, as well as self-supervised vision models such as DINOv2~\citep{oquab2023dinov2}.

More broadly, our results reveal systematic differences in the semantic decisions made by human and machine vision systems. These behavioral differences may reflect differences in their underlying visual representations, but our current experiments do not directly measure representational geometry. Our results show that perceptual decision boundaries do not necessarily align with the midpoint of CLIP interpolations, highlighting the need to characterize how distances in embedding space relate to human perceptual similarity. Future work could relate embedding geometry to perceptual transitions and use psychophysical measurements to refine representations toward better human alignment. Our framework demonstrates how controlled semantic ambiguity can be used to quantify human–machine perceptual alignment and probe its representational basis, bridging human perception, discriminative models, and generative systems.

{
    \small
    \bibliographystyle{ieeenat_fullname}
    \bibliography{bibliography}
}

\clearpage
\appendix

\begin{center}
{\Large \textbf{Supplementary Material}}
\end{center}

In the Supplementary Material, we include the following materials:
\begin{itemize}
\item An expanded related work discussion (Supplementary~\ref{sec:related_work}).
\item Details on our framework and its outputs (Supplementary~\ref{sec:supp_duck_rabbit}).
\item The elephant-rabbit condition (Supplementary~\ref{sec:supp_elephant_rabbit}).
\item Semantic continuum endpoint control analysis (Supplementary~\ref{sec:supp_endpoint}).
\end{itemize}

\section{Related Work}
\label{sec:related_work}

\paragraph{Ambiguity in Human Perception.}
Ambiguity is a fundamental property of human perception, reflecting sensory input that is noisy and compatible with multiple interpretations. The human visual system unconsciously performs probabilistic inference, integrating multiple pieces of sensory evidence with prior knowledge to attempt to resolve the uncertainty~\citep{knill2004bayesian}. We become aware of this inference when viewing a bistable or indeterminate image. \textit{Bistable images} yield two discrete interpretations that may alternate spontaneously despite a constant stimulus, such as in the classic duck–rabbit illusion~\citep{rock1983logic, leopold1999multistable}. Bistable images have been extensively studied using psychophysical methods that quantify dominance durations, switching dynamics, and perceptual bias~\citep{leopold1999multistable}. In contrast, \textit{indeterminate imagery} lacks sufficient perceptual evidence to support a coherent interpretation, resisting categorization and foregrounding perceptual limits~\citep{pepperell2006seeing, hertzmann2020visual}. While widely discussed in perceptual theory and art practice, indeterminate images have been less systematically studied with controlled psychophysics.

\paragraph{Generative Ambiguous Imagery.}
Early work on neural style transfer demonstrated that deep generative models can blend visual structures and stylistic features from multiple images, producing hybrid imagery that combines characteristics of different sources~\citep{gatys2016image}. Such methods highlighted how learned visual representations can interpolate between visual semantic concepts and generate perceptually mixed images. Related interactive systems such as Artbreeder~\citep{artbreeder} leverage generative adversarial networks, including StyleGAN~\citep{karras2019style} and BigGAN~\citep{brock2018large}, to allow users to explore and blend images in latent space. These approaches show that generative models can interpolate visual semantics, but they do not control ambiguity along a perceptual continuum.

More recent work has used generative models to construct images that support multiple interpretations while keeping low-level image statistics largely consistent~\citep{boger2025visual}. Diffusion Illusion~\citep{burgert2024diffusion} adopts an explicit optimization-based method using score distillation sampling (SDS)~\citep{poole2022dreamfusion} to align different prompts with different views. While effective, the reliance on SDS leads to long sampling times and often lower visual quality. In contrast, Visual Anagrams~\citep{geng2024visual} proposes a step-by-step multi-view denoising procedure that enforces prompt consistency across views during sampling. Illusion Diffusion~\citep{tancik2023illusion}, which generates rotation-based illusions by sampling from a latent diffusion model~\citep{rombach2022high}, is closer to a compact implementation of multi-view prompt consistency without explicit optimization. These generative illusion methods primarily generate multi-view illusions, where different geometric transformations reveal different semantic interpretations. In contrast, we introduce a framework for generating continuous semantic ambiguities within a single view, enabling psychophysical measurement of perceptual decision boundaries.

\paragraph{Comparison of Human and Machine Perception of Generative Imagery.}
A growing body of work compares human perception with the behavior of modern vision models, revealing both shared sensitivities and systematic divergences. Early studies showed that deep convolutional networks trained for object recognition can achieve high performance on benchmark datasets while relying on image features that differ substantially from those used by humans~\citep{geirhos2018imagenet}. Subsequent work has explored these differences using controlled visual stimuli, including adversarial perturbations~\citep{szegedy2013intriguing}, metamers generated from model representations~\citep{freeman2011metamers, feather2019metamers, harrington2023coco}, and images optimized to strongly activate individual neurons or classes~\citep{nguyen2015deep}. These approaches show that models can make confident predictions on images that humans find ambiguous or misleading.

Beyond individual stimuli, another line of work compares the categorical decision boundaries of humans and machines, often motivated by improving robustness and human alignment. Human and machine decision boundaries can be directly compared using psychophysical methods, with boundary-based classifiers such as SVMs better capturing human internal decision space than prototype-based approaches \citep{graf2006classification}.  CNN training consistently learns suboptimal decision boundaries, with SVMs learning boundaries closer to optimal on identical training data~\citep{richardson2021bayes}. ImageNet-trained CNNs also exhibit a strong texture bias in stark contrast to humans, revealing fundamentally different classification strategies even when accuracy is matched~\citep{geirhos2018imagenet}. Related work shows that modifying architectural assumptions, such as introducing biologically inspired foveated representations, can substantially alter the representational properties and robustness of vision models~\citep{deza2020emergent}. In addition, adversarial images that fool neural networks can sometimes transfer to human observers under brief presentation or controlled conditions~\citep{zhou2019humans}. Other studies have constructed image continua between categories to measure decision boundaries in humans and models~\citep{peterson2019human}. However, generating controlled ambiguous stimuli remains challenging, particularly when the goal is to compare perceptual decision boundaries across observers and models. Our work addresses this gap by using generative models to construct parametric continua of ambiguous imagery and measuring psychometric responses from both human observers and pretrained classifiers.
\clearpage
\newpage
\section{Details on Framework and Outputs}\label{sec:supp_duck_rabbit}

\subsection{Human Participants} We recruited 40 participants from the local university population (undergraduate students to faculty). 20 completed the rabbit-duck condition (16 male, 4 female; ages 21–52) and 20 completed the elephant-rabbit condition (8 male, 12 female; ages 22–36). Psychophysical experiments use a small number of participants over a larger number of trials in an effort to capture the internal biases and sensitivity patterns within rather than across individual observers.

All participants provided online informed consent in accordance with the Declaration of Helsinki. All procedures were approved by an Institutional Review Board. Participants used their own devices to complete the experiment, including mobile devices. One participant's data from the duck-rabbit condition was excluded due to unrealistic response patterns, leaving n = 19 for duck-rabbit. Trials with response times below 150 ms or above 5000 ms were also excluded, affecting fewer than 3$\%$ of any single participant's data.

\subsection{2AFC Task} 
\begin{figure*}[t!]
  \centering
  \includegraphics[width=\linewidth]{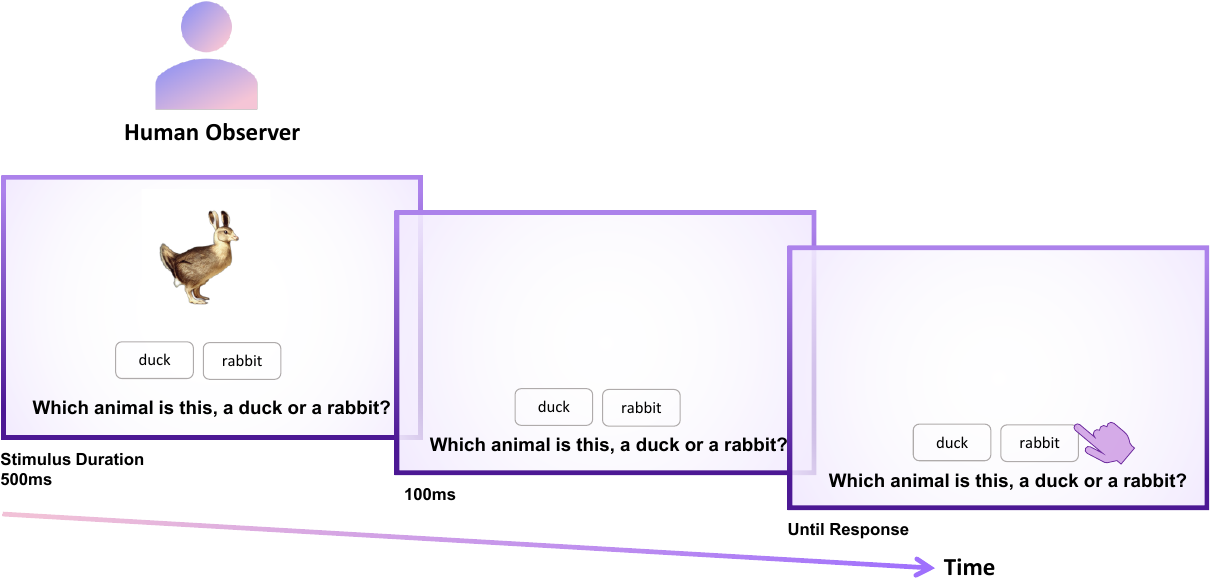}
 \caption{\textbf{Perceptual evaluation framework and 2AFC experimental design.} Participants performed a categorical two-alternative forced-choice task using ambiguous duck–rabbit images (300 trials per participant).} 
    \label{fig:preceptual_eval_framework}
\end{figure*}

For an overview of the perceptual evaluation framework and the 2AFC experimental design, see Fig.~\ref{fig:preceptual_eval_framework}. Participants completed a two-alternative forced-choice (2AFC) task, viewing one image per trial for 500 ms followed by a response screen prompting them to choose between the two semantic labels (rabbit or duck) as quickly and accurately as possible. The next trial began automatically upon response.  Each participant completed all 300 trials in random order, using the method of constant stimuli.

\subsection{Stimuli Generation Details}\label{sec:app_stimuli_generation}

\begin{figure*}[t]
    \centering
    \includegraphics[width=\linewidth]{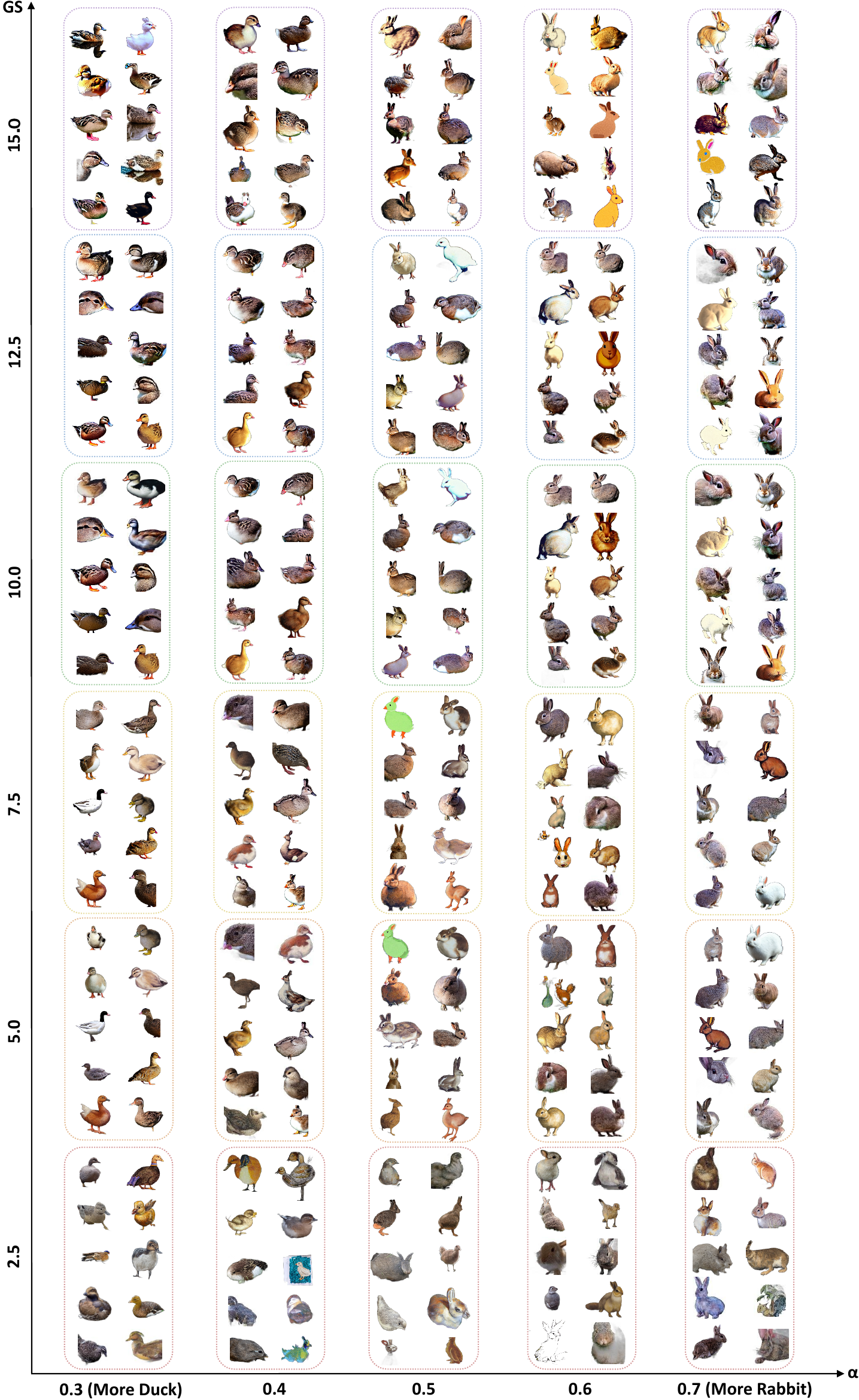}
    \caption{The 300 duck-rabbit stimuli used in the experiment, as a function of semantic mixing ratio ($\alpha$) and guidance scale (GS).}
    \label{fig:stimuli_generation_duck_rabbit}
\end{figure*}

\begin{figure*}[t]
    \centering
    \includegraphics[width=\linewidth]{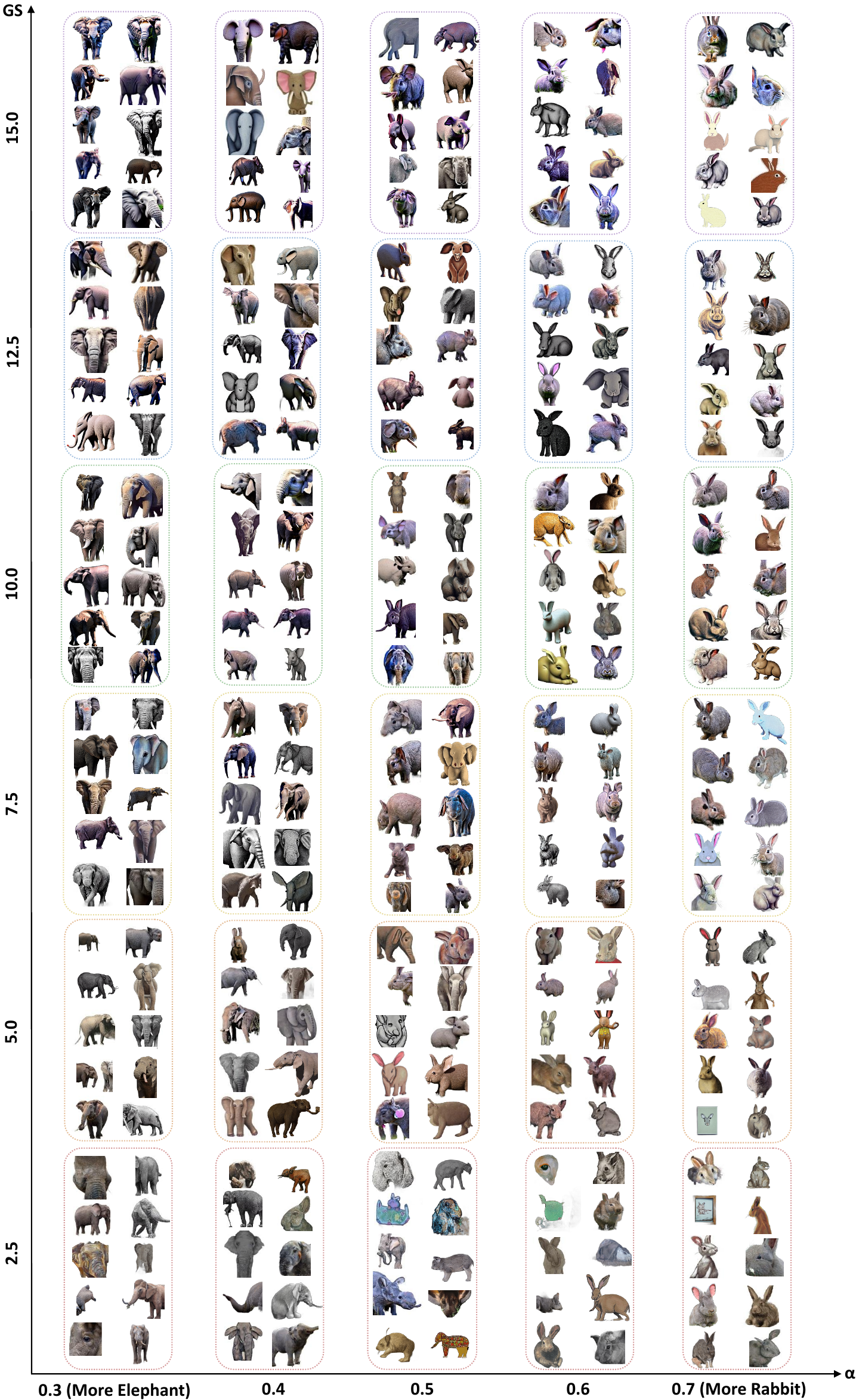}
    \caption{The 300 elephant-rabbit stimuli used in the experiment, as a function of semantic mixing ratio ($\alpha$) and guidance scale (GS).}
    \label{fig:stimuli_generation_elephant_rabbit}
\end{figure*}

Using our synthesis method, we generated 300 ambiguous stimuli for the duck-rabbit experiment $(5\,\alpha \ \text{levels} \times 6\,\text{guidance scales} \times 10\,\text{seeds})$. In Fig.~\ref{fig:stimuli_generation_duck_rabbit}, we present all 300 duck–rabbit stimuli used in the duck-rabbit experiment.
In addition to the duck-rabbit condition, we ran a condition with a second animal pair, creating ambiguous images between the prompts ``an elephant'' and ``a rabbit''. In Fig.~\ref{fig:stimuli_generation_elephant_rabbit}, we present all 300 elephant-rabbit stimuli used in the elephant-rabbit condition.

\clearpage
\newpage
\subsection{Classifier Models}\label{sec:supp_classifier_models}
We evaluated our ambiguous images on 8 pretrained ImageNet Classifiers ~\citep{russakovsky2015imagenet} spanning diverse architectures (Table~\ref{tab:model_overview}): ResNet-50~\citep{he2016deep}, ConvNeXt-Base~\citep{liu2022convnet}, DenseNet-121~\citep{huang2017densely}, MobileNetV3-L~\citep{howard2019searching}, EfficientNet-B0~\citep{tan2019efficientnet}, ViT-B/16~\citep{dosovitskiy2020image}, AlexNet~\citep{krizhevsky2012imagenet} and CORnet-S~\citep{kubilius2019brain}.

We chose a variety of ImageNet-1K classifier models with varying architectures, complexities, and training schemes, evaluating their performance on our duck-rabbit and elephant-rabbit classification tasks. We report the various model details in Table \ref{tab:model_overview}. To evaluate animal responses (duck/rabbit, elephant/rabbit) in models, we assigned the relevant ImageNet-1K labels to the associated animal category, taking the mean response over the relevant labels as the response probability for the animal category (Table \ref{tab:imagenet1k_class}).

\begin{table}[h]
\centering
\small
\resizebox{0.95\textwidth}{!}{
\begin{tabular}{llllcc}
\hline
\textbf{Category} &
\textbf{Model} &
\textbf{Architecture} &
\textbf{Training Objective} &
\textbf{Decision Space} \\
\hline

\multirow{8}{*}{\begin{tabular}[c]{@{}l@{}}Supervised learning \\ (ImageNet classification)\end{tabular}}
& ResNet-50~\citep{he2016deep}
& CNN
& ImageNet-1K classification
& 1000-way softmax \\
& ConvNeXt-Base~\citep{liu2022convnet}
& CNN
& ImageNet-1K classification
& 1000-way softmax \\
& DenseNet-121~\citep{huang2017densely}
& CNN
& ImageNet-1K classification
& 1000-way softmax \\
& MobileNetV3-L~\citep{howard2019searching}
& CNN
& ImageNet-1K classification
& 1000-way softmax \\
& EfficientNet-B0~\citep{tan2019efficientnet}
& CNN
& ImageNet-1K classification
& 1000-way softmax \\
& ViT-B/16~\citep{dosovitskiy2020image}
& Vision Transformer
& ImageNet-1K classification
& 1000-way softmax \\
& AlexNet~\citep{krizhevsky2012imagenet}
& CNN
& ImageNet-1K classification
& 1000-way softmax \\
& CORnet-S~\citep{kubilius2019brain}
& CNN
& ImageNet-1K classification
& 1000-way softmax \\
\hline
\end{tabular}
}
\vspace{0.6em}
\caption{\textbf{Models used for machine psychometric evaluation.} Eight classifiers are supervised models trained for ImageNet object recognition, and evaluated with a 1000-way softmax in the duck-rabbit categories. 
}
\label{tab:model_overview}
\end{table}
\begin{table}[h]
\centering
\resizebox{0.7\textwidth}{!}{
\begin{tabular}{c|l|l}
\hline
\textbf{Category} & \textbf{ImageNet-1K ID} & \textbf{Class Name} \\
\hline
\multirow{2}{*}{Duck (``a duck'')} 
\centering
& 97 & drake \\
& 98 & red-breasted merganser, \textit{Mergus serrator} \\
\hline
\multirow{3}{*}{Rabbit (``a rabbit'')} 
 & 330 & wood rabbit, cottontail, cottontail rabbit \\
 & 331 & hare \\
 & 332 & Angora, Angora rabbit \\
\hline
\multirow{2}{*}{Elephant (``an elephant'')} 
 & 385 & African elephant (\textit{Loxodonta africana}) \\
 & 386 & Indian elephant (\textit{Elephas maximus}) \\
\hline
\end{tabular}
}
\caption{Selected ImageNet-1k classes used for the rabbit, duck, and elephant.}
\label{tab:imagenet1k_class}
\end{table}

\subsection{Psychometric Analysis}\label{sec:app_psychophysics}

\textbf{Human observers.}
For each participant and guidance scale, we computed the proportion of ``rabbit'' responses at each of the five semantic mixing levels ($\alpha \in \{0.3,0.4,0.5,0.6,0.7\}$).
Psychometric functions were fitted using psignifit~\citep{schutt2016painfree} with a logistic sigmoid and equal lower and upper asymptote deviations, a standard approach in psychophysics~\citep{wichmann2001psychometric}:

$$
P(\text{rabbit}\mid\alpha)
= \lambda + (1-2\lambda)
\frac{1}{1+\exp[-\beta_1(\alpha-\mathrm{PSE})]},
$$

where $\lambda$ is the shared asymptote parameter, allowing fitted probabilities to approach $\lambda$ and $1-\lambda$ rather than 0 and 1, and $\beta_1$ controls the steepness of the sigmoid.  All responses were included in the fits.

The point of subjective equality (PSE) is the semantic mixing level at which $P(\text{rabbit}\mid\alpha)=0.5$. We defined response bias relative to the midpoint of the semantic continuum and perceptual sensitivity as the slope of the fitted function at the PSE:

$$
\begin{aligned}
\mathrm{Bias} &= \mathrm{PSE}-0.5,\\
\mathrm{Sensitivity}
&= \left.
\frac{\partial P(\text{rabbit}\mid\alpha)}
{\partial\alpha}
\right|_{\alpha=\mathrm{PSE}}
= \frac{(1-2\lambda)\beta_1}{4}.
\end{aligned}
$$

\begin{figure*}[h]
  \centering
  \includegraphics[width=\linewidth]{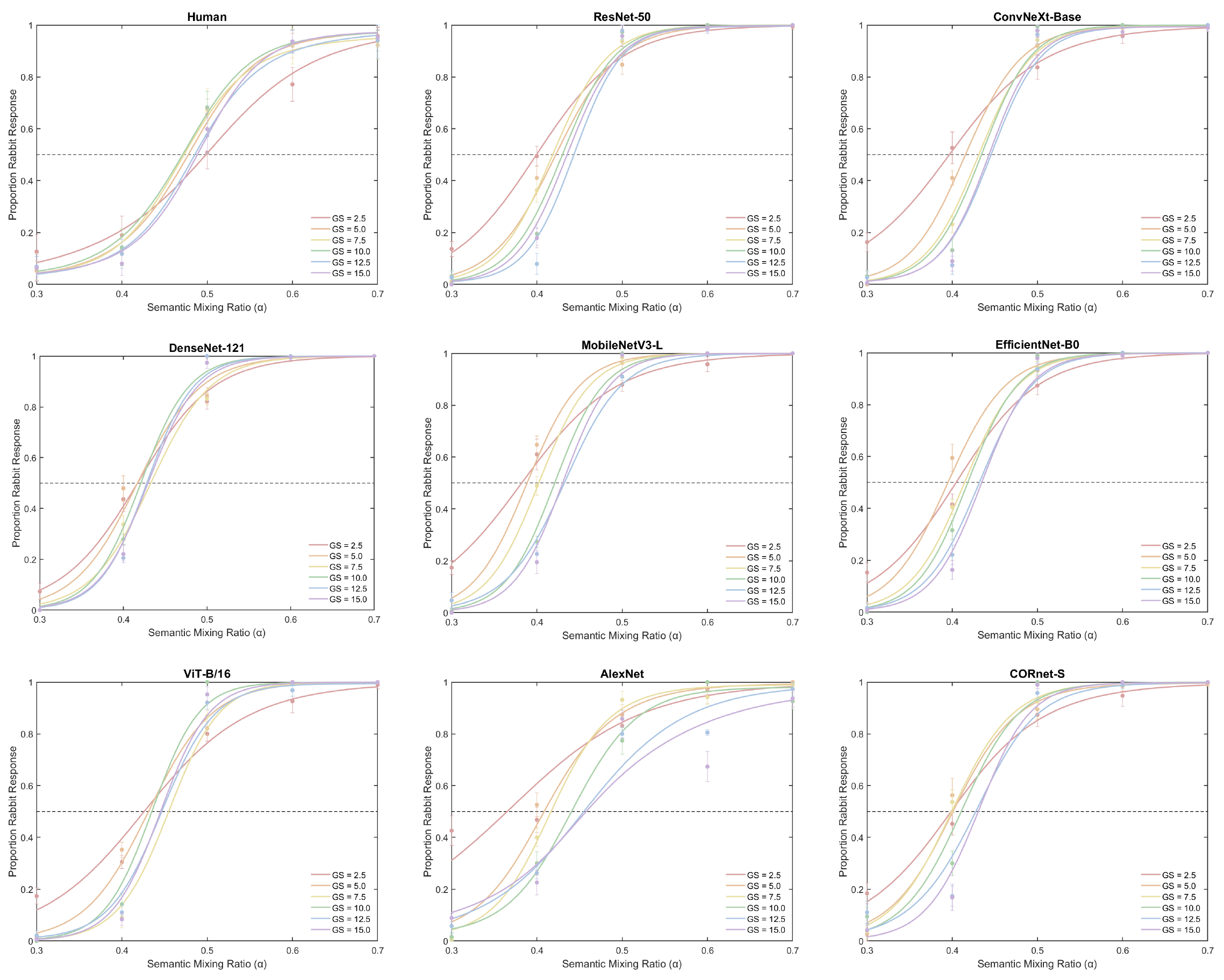}
  \vspace{-1em}
  \caption{
\textbf{Psychometric curves across guidance scales for humans and machine vision models in the duck-rabbit condition.} Points represent mean observed rabbit-response proportions, with error bars indicating 95\% confidence intervals (CIs) for 19 human participants and 19 simulated machine participants. Solid lines represent the mean of 19 individually fitted logistic psychometric functions for each guidance scale.}
\vspace{-1em}
    \label{fig:psychometric_curves_dr}
\end{figure*}

We averaged the bias and sensitivity across participants to create a grand average estimate for humans, with standard errors of the mean computed across individual participant estimates (Fig.~\ref{fig:psychometric_curves_dr}).

\textbf{Machine classifiers.}
We evaluated the same 300 duck–rabbit images using eight pretrained ImageNet classifiers (Table~\ref{tab:model_overview}). For each image + model combination, we extracted softmax probabilities over all ImageNet-1K classes and aggregated probabilities across the labels corresponding to each target animal. For animal $a$, the mean probability across its relevant ImageNet labels was

$$
\bar{p}_a =
\frac{1}{|\mathcal{A}_a|}
\sum_{i\in\mathcal{A}_a} p_i,
$$

where $\mathcal{A}_a$ is the set of ImageNet-1K labels corresponding to animal $a$, $p_i$ is the softmax probability assigned to label $i$, and $|\mathcal{A}_a|$ accounts for differences in the number of relevant ImageNet labels across animals (Table~\ref{tab:imagenet1k_class}). We normalized these aggregated probabilities to obtain

$$
p_{\mathrm{rabbit}} =
\frac{\bar{p}_{\text{rabbit}}}
{\bar{p}_{\text{rabbit}}+\bar{p}_{\text{duck}}}.
$$

For each machine classifier model, we simulated $n$ machine participants, where $n$ equals the number of human participants. For each simulated participant, we sampled each image response from a Bernoulli distribution of $p_{\mathrm{rabbit}}$. Analogous to the human participants, we fit each simulated machine participant using psignifit, and then averaged the bias and sensitivity over the $n$ simulated observers. While humans and machines are inherently different in their response formats, the method we present represents a principled and tractable approach to compare human and machine behavior as fairly as possible, though there are many other possible experimental designs. 

\textbf{Psychometric curves and goodness of psychometric fit for the duck-rabbit condition.}
Fig.~\ref{fig:psychometric_curves_dr} shows the psychometric functions and their corresponding goodness-of-fit scores in Table~\ref{tab:deviance_guidance_dr}. Goodness-of-fit was evaluated using the deviance statistics from psignifit, which computes the log-likelihood ratio between the fitted model and a saturated zero-residual model~\citep{wichmann2001psychometric}. All but one of the deviance scores fell below the critical chi-square value $\chi^2(2, 0.95) = 5.99$, indicating acceptable fits. AlexNet at GS = $15$ was just over at 6.01.

\begin{table}[h]
\centering
\resizebox{0.65\linewidth}{!}{
\begin{tabular}{c|cccccc}
\toprule
\diagbox{\textbf{Observer}}{\textbf{GS}}
& \textbf{2.5} & \textbf{5.0} & \textbf{7.5}
& \textbf{10.0} & \textbf{12.5} & \textbf{15.0} \\
\midrule
\textbf{Human}           & 3.05 & 1.98 & 2.69 & 2.84 & 3.19 & 2.82 \\
\textbf{ConvNeXt-Base}   & 1.56 & 1.44 & 1.40 & 3.62 & 4.43 & 4.39 \\
\textbf{DenseNet-121}   & 1.14 & 1.94 & 0.98 & 1.71 & 1.99 & 1.70 \\
\textbf{EfficientNet-B0} & 1.20 & 1.78 & 1.38 & 1.60 & 1.29 & 2.45 \\
\textbf{MobileNetV3-L}   & 1.34 & 1.81 & 1.38 & 1.84 & 1.35 & 2.10 \\
\textbf{ResNet-50}       & 0.98 & 1.44 & 1.12 & 2.61 & 4.33 & 2.21 \\
\textbf{ViT-B/16}        & 2.14 & 1.21 & 1.20 & 2.73 & 2.51 & 2.45 \\
\textbf{AlexNet}         & 3.43 & 2.58 & 2.76 & 2.92 & 3.07 & 6.01 \\
\textbf{CORnet-S}        & 2.12 & 2.42 & 1.10 & 3.15 & 4.55 & 3.43 \\
\bottomrule
\end{tabular}
}
\vspace{0.5em}
\caption{
\textbf{Deviance across guidance scales for human observers
and machine vision models in the duck-rabbit condition.}
Values represent mean deviance for 19 individual human
participants or 19 simulated machine participants.
Lower deviance values indicate better fits of the
psychometric function.
}
\label{tab:deviance_guidance_dr}
\end{table}

\begin{table*}[t!]
\centering
\begin{minipage}{\textwidth}
\centering
\setlength{\tabcolsep}{15pt}
\renewcommand{\arraystretch}{1.5}
\resizebox{\textwidth}{!}{%
\begin{tabular}{c|*{9}{c}}
\toprule
\diagbox{\textbf{GS}}{\textbf{Observer}}
& \headcell{Human}
& \headcell{ConvNeXt-Base}
& \headcell{DenseNet-121}
& \headcell{EfficientNet-B0}
& \headcell{MobileNetV3-L}
& \headcell{ResNet-50}
& \headcell{ViT-B/16}
& \headcell{AlexNet}
& \headcell{CORnet-S} \\
\midrule

\textbf{2.5}
& \eqcell{red!3}{\textbf{-0.006} \, [-0.025, 0.012]}
& \eqcell{red!45}{\textbf{-0.104} \, [-0.129, -0.097]}
& \eqcell{red!37}{\textbf{-0.084} \, [-0.101, -0.075]}
& \eqcell{red!42}{\textbf{-0.096} \, [-0.115, -0.086]}
& \eqcell{red!51}{\textbf{-0.118} \, [-0.145, -0.112]}
& \eqcell{red!43}{\textbf{-0.100} \, [-0.120, -0.091]}
& \eqcell{red!32}{\textbf{-0.073} \, [-0.093, -0.061]}
& \eqcell{red!60}{\textbf{-0.138} \, [-0.173, -0.131]}
& \eqcell{red!44}{\textbf{-0.101} \, [-0.124, -0.092]}
\\[0.25em]

\textbf{5.0}
& \eqcell{red!10}{\textbf{-0.022} \, [-0.038, -0.010]}
& \eqcell{red!36}{\textbf{-0.084} \, [-0.096, -0.073]}
& \eqcell{red!36}{\textbf{-0.083} \, [-0.098, -0.074]}
& \eqcell{red!45}{\textbf{-0.105} \, [-0.120, -0.096]}
& \eqcell{red!48}{\textbf{-0.111} \, [-0.125, -0.103]}
& \eqcell{red!34}{\textbf{-0.077} \, [-0.092, -0.068]}
& \eqcell{red!30}{\textbf{-0.069} \, [-0.082, -0.059]}
& \eqcell{red!39}{\textbf{-0.089} \, [-0.106, -0.080]}
& \eqcell{red!43}{\textbf{-0.099} \, [-0.115, -0.090]}
\\[0.25em]

\textbf{7.5}
& \eqcell{red!9}{\textbf{-0.020} \, [-0.036, -0.007]}
& \eqcell{red!30}{\textbf{-0.068} \, [-0.079, -0.059]}
& \eqcell{red!29}{\textbf{-0.067} \, [-0.081, -0.058]}
& \eqcell{red!37}{\textbf{-0.085} \, [-0.097, -0.076]}
& \eqcell{red!42}{\textbf{-0.098} \, [-0.109, -0.088]}
& \eqcell{red!35}{\textbf{-0.081} \, [-0.093, -0.072]}
& \eqcell{red!20}{\textbf{-0.045} \, [-0.056, -0.036]}
& \eqcell{red!35}{\textbf{-0.082} \, [-0.096, -0.072]}
& \eqcell{red!44}{\textbf{-0.101} \, [-0.116, -0.092]}
\\[0.25em]

\textbf{10.0}
& \eqcell{red!13}{\textbf{-0.031} \, [-0.044, -0.016]}
& \eqcell{red!29}{\textbf{-0.066} \, [-0.077, -0.056]}
& \eqcell{red!34}{\textbf{-0.079} \, [-0.089, -0.069]}
& \eqcell{red!35}{\textbf{-0.082} \, [-0.092, -0.072]}
& \eqcell{red!34}{\textbf{-0.079} \, [-0.090, -0.070]}
& \eqcell{red!31}{\textbf{-0.070} \, [-0.081, -0.061]}
& \eqcell{red!28}{\textbf{-0.065} \, [-0.076, -0.056]}
& \eqcell{red!26}{\textbf{-0.059} \, [-0.072, -0.046]}
& \eqcell{red!39}{\textbf{-0.090} \, [-0.104, -0.079]}
\\[0.25em]

\textbf{12.5}
& \eqcell{red!5}{\textbf{-0.011} \, [-0.028, 0.003]}
& \eqcell{red!24}{\textbf{-0.054} \, [-0.066, -0.044]}
& \eqcell{red!31}{\textbf{-0.072} \, [-0.082, -0.062]}
& \eqcell{red!30}{\textbf{-0.068} \, [-0.079, -0.058]}
& \eqcell{red!30}{\textbf{-0.069} \, [-0.081, -0.057]}
& \eqcell{red!25}{\textbf{-0.057} \, [-0.068, -0.047]}
& \eqcell{red!23}{\textbf{-0.053} \, [-0.065, -0.043]}
& \eqcell{red!20}{\textbf{-0.045} \, [-0.065, -0.036]}
& \eqcell{red!32}{\textbf{-0.074} \, [-0.087, -0.062]}
\\[0.25em]

\textbf{15.0}
& \eqcell{red!7}{\textbf{-0.015} \, [-0.030, -0.003]}
& \eqcell{red!25}{\textbf{-0.057} \, [-0.067, -0.046]}
& \eqcell{red!31}{\textbf{-0.070} \, [-0.080, -0.061]}
& \eqcell{red!28}{\textbf{-0.065} \, [-0.076, -0.055]}
& \eqcell{red!30}{\textbf{-0.070} \, [-0.081, -0.061]}
& \eqcell{red!28}{\textbf{-0.064} \, [-0.075, -0.054]}
& \eqcell{red!24}{\textbf{-0.055} \, [-0.065, -0.045]}
& \eqcell{red!16}{\textbf{-0.036} \, [-0.059, -0.023]}
& \eqcell{red!30}{\textbf{-0.070} \, [-0.081, -0.060]}
\\

\bottomrule
\end{tabular}
}

\caption{
\textbf{Bias estimates for human observers and machine vision models across guidance scales.}
Bias is defined as $\mathrm{PSE}-0.5$.
Each cell reports the mean bias estimate and its 95\% confidence interval (CI), in the format
$\mathrm{Bias}\,[\mathrm{lower},\mathrm{upper}]$.
White indicates zero bias, red indicates negative bias (toward rabbits), and blue indicates positive bias (toward ducks).
Color intensity is proportional to the absolute bias estimate, normalized to the maximum absolute bias across all entries, with the maximum mapped to 60\% color intensity.
}

\label{tab:bias_heat}
\end{minipage}
\end{table*}

\begin{table*}[t!]
\centering
\begingroup
\setlength{\tabcolsep}{15pt}
\renewcommand{\arraystretch}{1.5}
\resizebox{\textwidth}{!}{%
\begin{tabular}{c|*{9}{c}}
\toprule
\diagbox{\textbf{GS}}{\textbf{Observer}}
& \headcell{Human}
& \headcell{ConvNeXt-Base}
& \headcell{DenseNet-121}
& \headcell{EfficientNet-B0}
& \headcell{MobileNetV3-L}
& \headcell{ResNet-50}
& \headcell{ViT-B/16}
& \headcell{AlexNet}
& \headcell{CORnet-S}
\\
\midrule
\textbf{2.5}
& \eqcell{red!26}{\textbf{3.94} \, [3.042, 4.194]}
& \eqcell{red!30}{\textbf{4.63} \, [3.464, 4.756]}
& \eqcell{red!37}{\textbf{5.65} \, [4.177, 5.610]}
& \eqcell{red!34}{\textbf{5.26} \, [3.885, 5.288]}
& \eqcell{red!30}{\textbf{4.56} \, [3.477, 4.819]}
& \eqcell{red!33}{\textbf{5.11} \, [3.777, 5.135]}
& \eqcell{red!27}{\textbf{4.10} \, [3.287, 4.535]}
& \eqcell{red!21}{\textbf{3.19} \, [2.488, 3.421]}
& \eqcell{red!30}{\textbf{4.66} \, [3.442, 4.766]}
\\[0.25em]
\textbf{5.0}
& \eqcell{red!40}{\textbf{6.20} \, [4.385, 5.886]}
& \eqcell{red!50}{\textbf{7.68} \, [5.420, 7.052]}
& \eqcell{red!44}{\textbf{6.77} \, [4.823, 6.424]}
& \eqcell{red!49}{\textbf{7.57} \, [5.237, 6.944]}
& \eqcell{red!53}{\textbf{8.07} \, [5.693, 7.424]}
& \eqcell{red!45}{\textbf{6.89} \, [4.901, 6.430]}
& \eqcell{red!44}{\textbf{6.72} \, [4.843, 6.382]}
& \eqcell{red!40}{\textbf{6.21} \, [4.442, 6.033]}
& \eqcell{red!45}{\textbf{6.84} \, [4.762, 6.413]}
\\[0.25em]
\textbf{7.5}
& \eqcell{red!42}{\textbf{6.37} \, [4.379, 5.904]}
& \eqcell{red!53}{\textbf{8.16} \, [5.997, 7.731]}
& \eqcell{red!46}{\textbf{7.09} \, [5.101, 6.635]}
& \eqcell{red!51}{\textbf{7.89} \, [5.631, 7.246]}
& \eqcell{red!54}{\textbf{8.30} \, [5.972, 7.736]}
& \eqcell{red!52}{\textbf{7.92} \, [5.741, 7.400]}
& \eqcell{red!56}{\textbf{8.54} \, [6.378, 8.120]}
& \eqcell{red!47}{\textbf{7.23} \, [4.989, 6.652]}
& \eqcell{red!47}{\textbf{7.23} \, [5.100, 6.713]}
\\[0.25em]
\textbf{10.0}
& \eqcell{red!41}{\textbf{6.35} \, [4.497, 6.004]}
& \eqcell{red!57}{\textbf{8.75} \, [6.465, 8.284]}
& \eqcell{red!57}{\textbf{8.69} \, [6.447, 8.153]}
& \eqcell{red!56}{\textbf{8.55} \, [6.285, 8.037]}
& \eqcell{red!56}{\textbf{8.63} \, [6.319, 8.112]}
& \eqcell{red!55}{\textbf{8.39} \, [6.142, 7.867]}
& \eqcell{red!60}{\textbf{9.20} \, [6.890, 8.633]}
& \eqcell{red!42}{\textbf{6.47} \, [4.459, 6.033]}
& \eqcell{red!49}{\textbf{7.46} \, [5.349, 7.028]}
\\[0.25em]
\textbf{12.5}
& \eqcell{red!40}{\textbf{6.08} \, [4.354, 5.866]}
& \eqcell{red!57}{\textbf{8.81} \, [6.360, 8.155]}
& \eqcell{red!58}{\textbf{8.88} \, [6.618, 8.390]}
& \eqcell{red!53}{\textbf{8.09} \, [5.959, 7.676]}
& \eqcell{red!48}{\textbf{7.40} \, [5.431, 7.049]}
& \eqcell{red!59}{\textbf{9.05} \, [6.614, 8.345]}
& \eqcell{red!54}{\textbf{8.21} \, [5.956, 7.710]}
& \eqcell{red!27}{\textbf{4.17} \, [3.265, 4.612]}
& \eqcell{red!45}{\textbf{6.92} \, [5.203, 6.934]}
\\[0.25em]
\textbf{15.0}
& \eqcell{red!44}{\textbf{6.79} \, [4.701, 6.272]}
& \eqcell{red!59}{\textbf{9.02} \, [6.519, 8.321]}
& \eqcell{red!56}{\textbf{8.61} \, [6.348, 8.115]}
& \eqcell{red!57}{\textbf{8.75} \, [6.512, 8.200]}
& \eqcell{red!58}{\textbf{8.93} \, [6.682, 8.384]}
& \eqcell{red!56}{\textbf{8.63} \, [6.310, 8.043]}
& \eqcell{red!59}{\textbf{9.03} \, [6.717, 8.515]}
& \eqcell{red!26}{\textbf{3.92} \, [2.850, 4.204]}
& \eqcell{red!56}{\textbf{8.52} \, [6.258, 8.083]}
\\
\bottomrule
\end{tabular}%
}
\caption{
\textbf{Sensitivity estimates for human observers and machine vision models across guidance scales.}
Sensitivity is defined as the slope of the fitted psychometric function at the point of subjective equality (PSE).
Each cell reports the estimated slope and its reported 95\% confidence interval (CI) in the format
$\mathrm{Slope}\,[\mathrm{lower},\mathrm{upper}]$.
Darker red indicates greater estimated sensitivity; color intensity is proportional to the point estimate,
normalized to the maximum slope across all entries (mapped to 60\% color intensity).
}
\label{tab:slope}
\endgroup
\end{table*}

\begin{figure*}[t!]
  \centering
  \includegraphics[width=\linewidth]{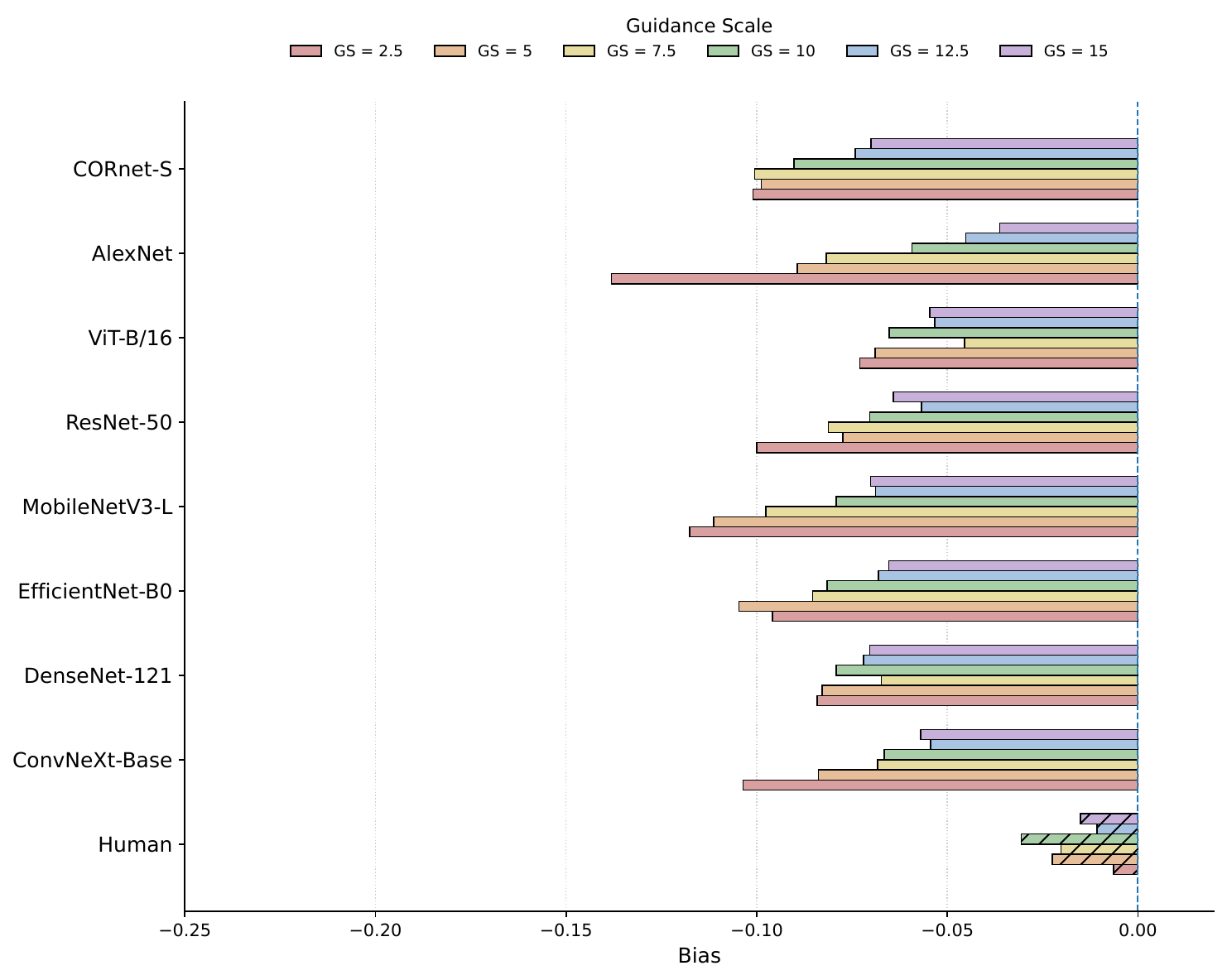} 
 \caption{\textbf{Bias for the duck-rabbit condition across guidance scales for humans and classifier models.}
Grouped bars show bias values computed as $\mathrm{PSE}-0.5$ for each model, with colors indicating different guidance scales (GS). Bias was negative for all humans and machine classifiers, indicating a consistent bias towards rabbit.
}
    \label{fig:bias_gs_bar}
\end{figure*}

\begin{figure*}[t!]
  \centering
  \includegraphics[width=\linewidth]{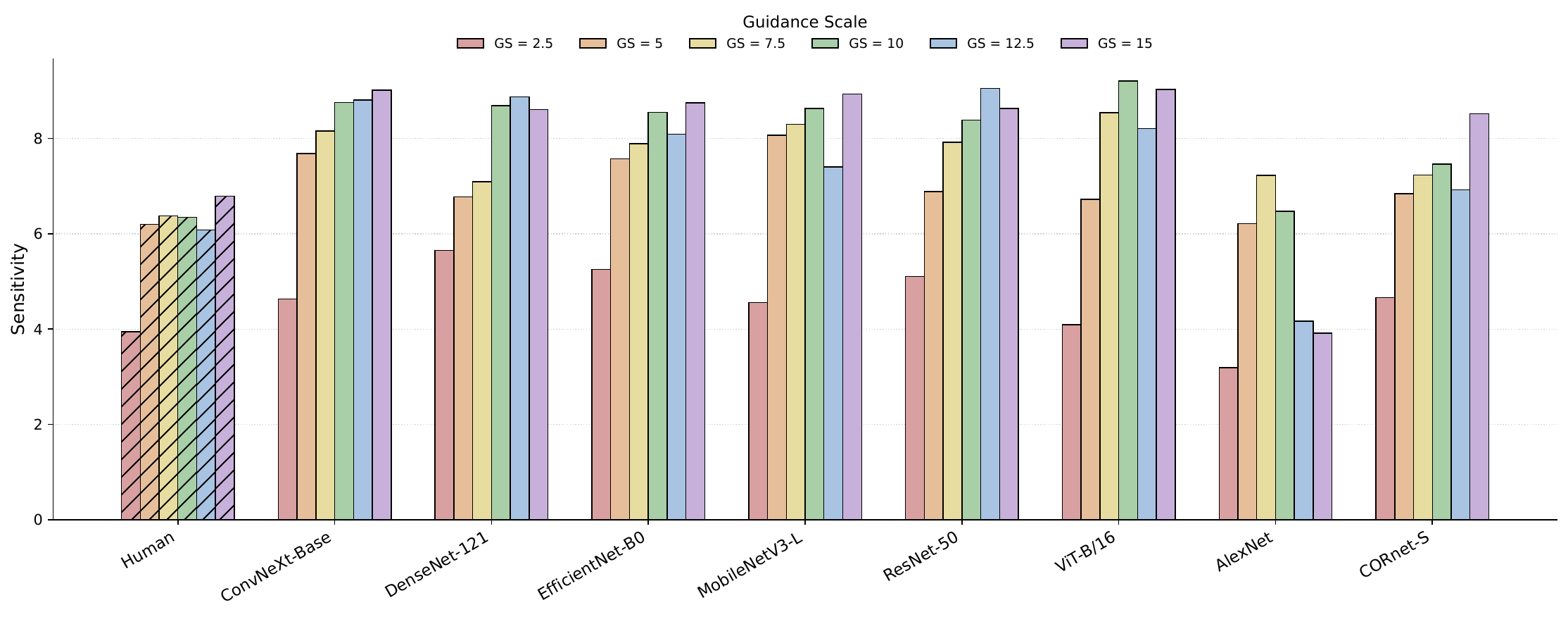}
\caption{\textbf{Sensitivity for the duck-rabbit condition across guidance scales for humans and classifier models.}
Higher sensitivity values indicate stronger semantic separation as the mixing parameter $\alpha$ varies.
Grouped bars show sensitivity values for each model, with colors indicating guidance scales (GS).
Sensitivity generally increases with guidance scale for both humans and models, although the magnitude and stability of this increase vary across architectures.
}
\label{fig:sensitivity_gs_bar}
\end{figure*}

\newpage
\clearpage


\section{Elephant-Rabbit Condition}
\label{sec:supp_elephant_rabbit}

We apply a parallel pipeline to our duck-rabbit condition on a second ambiguous animal pairing: elephants and rabbits. This second pair serves two purposes. First, it tests whether the observed human–model differences generalize beyond a canonical bistable-inspired example. Second, it introduces a pairing with greater visual and semantic dissimilarity (e.g., large vs. small, trunk vs. ears), allowing us to probe whether bias and sensitivity depend on the structure of the underlying categories rather than the specific illusion.

Stimuli were generated by interpolating between the CLIP embeddings corresponding to ``an elephant'' and ``a rabbit''. Human observers completed the corresponding 2AFC task, and the same pretrained ImageNet classifiers were evaluated on the generated stimuli. Human and machine responses were analyzed using the same psychometric procedure described above. 

\begin{figure*}[h]
  \centering
  \vspace{-1.5em}
  \includegraphics[width=0.98\linewidth]{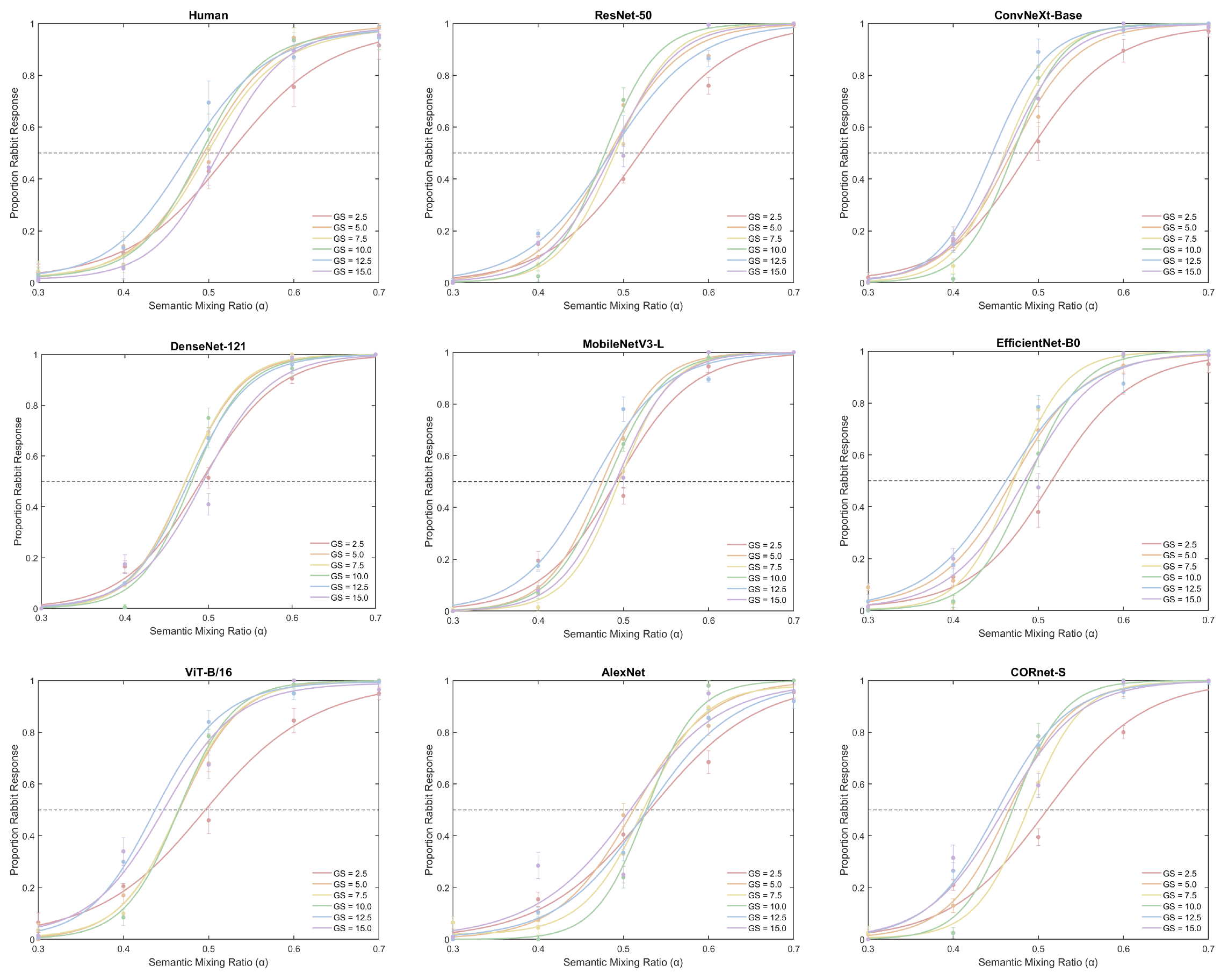}
  \vspace{-0.5em}
  \caption{
\textbf{Psychometric curves across guidance scales in the elephant-rabbit condition.} Points represent mean observed rabbit-response proportions, with error bars indicating 95\% confidence intervals (CIs) for 20 human participants or 20 simulated machine participants. Solid lines represent the mean of 20 individually fitted logistic psychometric functions for the guidance scale.}
    \label{fig:psychometric_curves_er}
\end{figure*}

\textbf{Psychometric curves and goodness of psychometric fit for the elephant-rabbit condition.} Fig.~\ref{fig:psychometric_curves_er} shows the psychometric functions and their corresponding goodness-of-fit scores in Table~\ref{tab:deviance_guidance_er}. All but
one of the deviance scores fell below the critical chi-square value $\chi^2(2, 0.95) = 5.99$, indicating
acceptable fits. Again, AlexNet at GS = $15$ was just over at 6.84.

\begin{table}[h]
\centering
\resizebox{0.6\linewidth}{!}{
\begin{tabular}{c|cccccc}
\toprule
\diagbox{\textbf{Observer}}{\textbf{GS}}
& \textbf{2.5} & \textbf{5.0} & \textbf{7.5}
& \textbf{10.0} & \textbf{12.5} & \textbf{15.0} \\
\midrule
\textbf{Human}           & 2.61 & 2.22 & 2.18 & 2.50 & 2.79 & 2.32 \\
\textbf{ConvNeXt-Base}   & 2.22 & 1.35 & 2.17 & 1.77 & 2.16 & 1.31 \\
\textbf{DenseNet-121}    & 1.07 & 0.60 & 0.61 & 2.71 & 0.94 & 2.98 \\
\textbf{EfficientNet-B0} & 2.35 & 2.28 & 2.17 & 1.26 & 2.90 & 3.04 \\
\textbf{MobileNetV3-L}   & 2.21 & 0.81 & 1.47 & 1.07 & 1.91 & 1.29 \\
\textbf{ResNet-50}       & 1.70 & 1.43 & 1.38 & 1.50 & 1.72 & 1.98 \\
\textbf{ViT-B/16}        & 2.16 & 1.03 & 1.49 & 1.38 & 2.03 & 3.13 \\
\textbf{AlexNet}         & 2.29 & 1.42 & 2.01 & 1.61 & 1.69 & 6.84 \\
\textbf{CORnet-S}        & 2.08 & 1.18 & 1.97 & 1.79 & 1.54 & 2.81 \\
\bottomrule
\end{tabular}
}
\vspace{0.5em}
\caption{
\textbf{Mean deviance across guidance scales for human observers and machine vision models in the elephant-rabbit condition.}
Values represent mean deviance for 20 individual participants for humans and 20 simulated machine participants. Lower deviance values indicate better fits of the psychometric function.
}
\label{tab:deviance_guidance_er}
\end{table}

\begin{table*}[t]
\centering
\begin{minipage}{\textwidth}
\centering
\setlength{\tabcolsep}{15pt}
\renewcommand{\arraystretch}{1.5}
\resizebox{\textwidth}{!}{%
\begin{tabular}{c|*{9}{c}}
\toprule
\diagbox{\textbf{GS}}{\textbf{Observer}}
& \headcell{Human}
& \headcell{ConvNeXt-Base}
& \headcell{DenseNet-121}
& \headcell{EfficientNet-B0}
& \headcell{MobileNetV3-L}
& \headcell{ResNet-50}
& \headcell{ViT-B/16}
& \headcell{AlexNet}
& \headcell{CORnet-S} \\
\midrule

\textbf{2.5}
& \eqcell{blue!26}{\textbf{0.026} \, [0.013, 0.044]}
& \eqcell{red!9}{\textbf{-0.009} \, [-0.022, 0.003]}
& \eqcell{red!8}{\textbf{-0.008} \, [-0.020, 0.004]}
& \eqcell{blue!16}{\textbf{0.017} \, [0.004, 0.029]}
& \eqcell{red!8}{\textbf{-0.008} \, [-0.020, 0.004]}
& \eqcell{blue!19}{\textbf{0.020} \, [0.008, 0.035]}
& \eqcell{red!3}{\textbf{-0.003} \, [-0.017, 0.013]}
& \eqcell{blue!31}{\textbf{0.031} \, [0.019, 0.049]}
& \eqcell{blue!10}{\textbf{0.010} \, [-0.001, 0.025]} \\[0.25em]

\textbf{5.0}
& \eqcell{red!6}{\textbf{-0.006} \, [-0.018, 0.007]}
& \eqcell{red!31}{\textbf{-0.031} \, [-0.043, -0.020]}
& \eqcell{red!27}{\textbf{-0.027} \, [-0.039, -0.018]}
& \eqcell{red!31}{\textbf{-0.031} \, [-0.045, -0.021]}
& \eqcell{red!24}{\textbf{-0.025} \, [-0.035, -0.015]}
& \eqcell{red!16}{\textbf{-0.016} \, [-0.030, -0.007]}
& \eqcell{red!34}{\textbf{-0.035} \, [-0.046, -0.025]}
& \eqcell{blue!12}{\textbf{0.012} \, [-0.000, 0.024]}
& \eqcell{red!34}{\textbf{-0.035} \, [-0.047, -0.025]} \\[0.25em]

\textbf{7.5}
& \eqcell{blue!1}{\textbf{0.001} \, [-0.012, 0.014]}
& \eqcell{red!38}{\textbf{-0.039} \, [-0.050, -0.030]}
& \eqcell{red!27}{\textbf{-0.027} \, [-0.038, -0.018]}
& \eqcell{red!28}{\textbf{-0.029} \, [-0.040, -0.020]}
& \eqcell{red!5}{\textbf{-0.005} \, [-0.016, 0.003]}
& \eqcell{red!9}{\textbf{-0.010} \, [-0.019, 0.001]}
& \eqcell{red!35}{\textbf{-0.036} \, [-0.048, -0.027]}
& \eqcell{blue!22}{\textbf{0.023} \, [0.011, 0.034]}
& \eqcell{red!12}{\textbf{-0.013} \, [-0.025, -0.004]} \\[0.25em]

\textbf{10.0}
& \eqcell{red!6}{\textbf{-0.006} \, [-0.019, 0.006]}
& \eqcell{red!29}{\textbf{-0.029} \, [-0.039, -0.020]}
& \eqcell{red!19}{\textbf{-0.019} \, [-0.030, -0.011]}
& \eqcell{red!12}{\textbf{-0.012} \, [-0.022, -0.002]}
& \eqcell{red!18}{\textbf{-0.019} \, [-0.029, -0.009]}
& \eqcell{red!21}{\textbf{-0.022} \, [-0.032, -0.013]}
& \eqcell{red!34}{\textbf{-0.035} \, [-0.046, -0.026]}
& \eqcell{blue!26}{\textbf{0.027} \, [0.018, 0.037]}
& \eqcell{red!29}{\textbf{-0.030} \, [-0.039, -0.021]} \\[0.25em]

\textbf{12.5}
& \eqcell{red!18}{\textbf{-0.019} \, [-0.034, -0.007]}
& \eqcell{red!52}{\textbf{-0.053} \, [-0.064, -0.043]}
& \eqcell{red!22}{\textbf{-0.023} \, [-0.034, -0.013]}
& \eqcell{red!37}{\textbf{-0.038} \, [-0.054, -0.028]}
& \eqcell{red!35}{\textbf{-0.036} \, [-0.050, -0.028]}
& \eqcell{red!14}{\textbf{-0.015} \, [-0.029, -0.003]}
& \eqcell{red!60}{\textbf{-0.061} \, [-0.075, -0.052]}
& \eqcell{blue!28}{\textbf{0.029} \, [0.016, 0.042]}
& \eqcell{red!46}{\textbf{-0.047} \, [-0.061, -0.037]} \\[0.25em]

\textbf{15.0}
& \eqcell{blue!13}{\textbf{0.013} \, [0.002, 0.026]}
& \eqcell{red!37}{\textbf{-0.038} \, [-0.050, -0.027]}
& \eqcell{red!7}{\textbf{-0.007} \, [-0.018, 0.006]}
& \eqcell{red!17}{\textbf{-0.018} \, [-0.028, -0.004]}
& \eqcell{red!9}{\textbf{-0.009} \, [-0.019, 0.001]}
& \eqcell{red!14}{\textbf{-0.014} \, [-0.025, -0.002]}
& \eqcell{red!52}{\textbf{-0.054} \, [-0.068, -0.042]}
& \eqcell{blue!6}{\textbf{0.006} \, [-0.007, 0.023]}
& \eqcell{red!40}{\textbf{-0.041} \, [-0.054, -0.030]} \\

\bottomrule
\end{tabular}
}
\caption{
\textbf{Bias estimates for human observers and machine vision models in the elephant--rabbit condition.}
Bias is defined as $\mathrm{PSE}-0.5$.
Each cell reports the bias estimate and its reported 95\% confidence interval (CI)
in the format $\mathrm{Bias}\,[\mathrm{lower},\mathrm{upper}]$.
The reported interval bounds are shifted by $-0.5$ from the PSE scale.
Negative bias (red) indicates a tendency toward rabbits, and positive bias (blue)
indicates a tendency toward elephants. White denotes zero bias.
Color intensity is proportional to the absolute point-estimate bias, normalized
to the maximum absolute bias across all entries, with the maximum mapped to 60\% color intensity.
}
\label{tab:bias_er}
\end{minipage}
\end{table*}
\begin{table*}[t]
\centering
\begin{minipage}{\textwidth}
\centering
\setlength{\tabcolsep}{15pt}
\renewcommand{\arraystretch}{1.5}
\resizebox{\textwidth}{!}{%
\begin{tabular}{c|*{9}{c}}
\toprule
\diagbox{\textbf{GS}}{\textbf{Observer}}
& \headcell{Human}
& \headcell{ConvNeXt-Base}
& \headcell{DenseNet-121}
& \headcell{EfficientNet-B0}
& \headcell{MobileNetV3-L}
& \headcell{ResNet-50}
& \headcell{ViT-B/16}
& \headcell{AlexNet}
& \headcell{CORnet-S} \\
\midrule
\textbf{2.5}
& \eqcell{red!16}{\textbf{4.55} \, [3.548, 4.781]}
& \eqcell{red!24}{\textbf{5.32} \, [4.104, 5.490]}
& \eqcell{red!26}{\textbf{5.54} \, [4.287, 5.669]}
& \eqcell{red!25}{\textbf{5.40} \, [4.141, 5.540]}
& \eqcell{red!28}{\textbf{5.73} \, [4.399, 5.720]}
& \eqcell{red!17}{\textbf{4.68} \, [3.674, 4.893]}
& \eqcell{red!10}{\textbf{3.98} \, [3.284, 4.454]}
& \eqcell{red!10}{\textbf{4.02} \, [3.221, 4.344]}
& \eqcell{red!15}{\textbf{4.44} \, [3.519, 4.696]}
\\[0.25em]
\textbf{5.0}
& \eqcell{red!32}{\textbf{6.12} \, [4.549, 6.006]}
& \eqcell{red!38}{\textbf{6.63} \, [4.863, 6.313]}
& \eqcell{red!47}{\textbf{7.51} \, [5.527, 7.139]}
& \eqcell{red!30}{\textbf{5.90} \, [4.429, 5.912]}
& \eqcell{red!50}{\textbf{7.80} \, [5.722, 7.282]}
& \eqcell{red!30}{\textbf{5.91} \, [4.605, 6.042]}
& \eqcell{red!43}{\textbf{7.11} \, [5.259, 6.724]}
& \eqcell{red!29}{\textbf{5.76} \, [4.379, 5.749]}
& \eqcell{red!41}{\textbf{6.94} \, [5.174, 6.666]}
\\[0.25em]
\textbf{7.5}
& \eqcell{red!34}{\textbf{6.28} \, [4.607, 6.104]}
& \eqcell{red!57}{\textbf{8.44} \, [6.265, 7.924]}
& \eqcell{red!50}{\textbf{7.77} \, [5.678, 7.240]}
& \eqcell{red!57}{\textbf{8.43} \, [6.254, 7.931]}
& \eqcell{red!55}{\textbf{8.32} \, [6.137, 7.793]}
& \eqcell{red!50}{\textbf{7.80} \, [5.713, 7.284]}
& \eqcell{red!45}{\textbf{7.34} \, [5.338, 6.958]}
& \eqcell{red!41}{\textbf{6.96} \, [4.844, 6.327]}
& \eqcell{red!54}{\textbf{8.20} \, [5.979, 7.623]}
\\[0.25em]
\textbf{10.0}
& \eqcell{red!42}{\textbf{7.04} \, [5.001, 6.527]}
& \eqcell{red!60}{\textbf{8.76} \, [6.621, 8.297]}
& \eqcell{red!53}{\textbf{8.05} \, [6.045, 7.703]}
& \eqcell{red!52}{\textbf{8.04} \, [5.966, 7.550]}
& \eqcell{red!49}{\textbf{7.68} \, [5.631, 7.242]}
& \eqcell{red!56}{\textbf{8.40} \, [6.337, 7.886]}
& \eqcell{red!51}{\textbf{7.92} \, [5.906, 7.494]}
& \eqcell{red!59}{\textbf{8.66} \, [6.476, 8.199]}
& \eqcell{red!60}{\textbf{8.72} \, [6.570, 8.231]}
\\[0.25em]
\textbf{12.5}
& \eqcell{red!31}{\textbf{5.98} \, [4.377, 5.829]}
& \eqcell{red!52}{\textbf{7.96} \, [5.940, 7.539]}
& \eqcell{red!43}{\textbf{7.15} \, [5.255, 6.832]}
& \eqcell{red!25}{\textbf{5.37} \, [4.281, 5.784]}
& \eqcell{red!30}{\textbf{5.93} \, [4.649, 6.126]}
& \eqcell{red!21}{\textbf{5.05} \, [3.992, 5.238]}
& \eqcell{red!37}{\textbf{6.57} \, [4.814, 6.342]}
& \eqcell{red!19}{\textbf{4.88} \, [3.987, 5.362]}
& \eqcell{red!33}{\textbf{6.17} \, [4.589, 6.058]}
\\[0.25em]
\textbf{15.0}
& \eqcell{red!38}{\textbf{6.62} \, [4.735, 6.233]}
& \eqcell{red!43}{\textbf{7.18} \, [5.230, 6.747]}
& \eqcell{red!35}{\textbf{6.35} \, [4.744, 6.215]}
& \eqcell{red!31}{\textbf{6.03} \, [4.460, 5.932]}
& \eqcell{red!49}{\textbf{7.71} \, [5.579, 7.118]}
& \eqcell{red!39}{\textbf{6.79} \, [4.926, 6.469]}
& \eqcell{red!30}{\textbf{5.89} \, [4.247, 5.613]}
& \eqcell{red!18}{\textbf{4.73} \, [3.796, 5.141]}
& \eqcell{red!29}{\textbf{5.81} \, [4.329, 5.678]}
\\
\bottomrule
\end{tabular}%
}
\caption{\textbf{Sensitivity estimates for human observers and classifier models across guidance scales in the elephant--rabbit condition.}
Sensitivity is defined as the slope of the fitted logistic psychometric function at the point of subjective equality (PSE). 
Each cell reports the mean fitted sensitivity and its reported 95\% confidence interval (CI), in the format $\mathrm{Sensitivity}\,[\mathrm{lower},\mathrm{upper}]$. Darker red indicates higher estimated sensitivity. Color intensity is linearly normalized from 10\% to 60\% according to point estimates across all entries, independently of CI width.
}
\label{tab:slope_er}
\end{minipage}
\end{table*}

\begin{figure*}[h]
  \centering
  \vspace{-0.8em}
  \includegraphics[width=\linewidth]{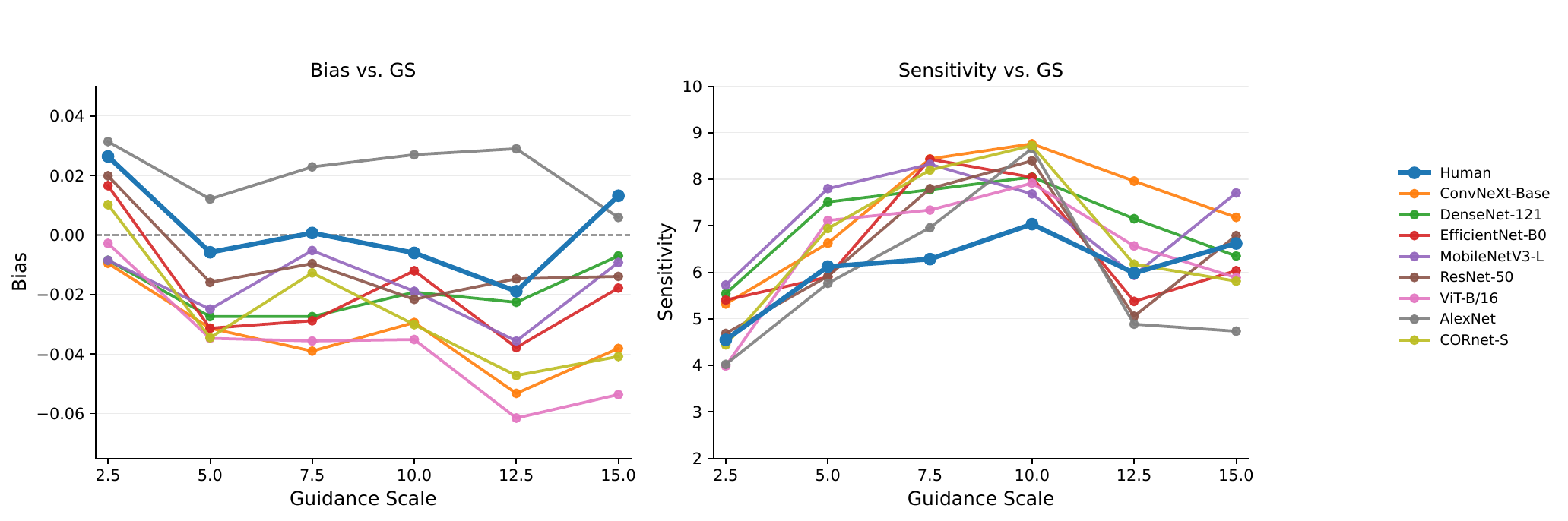}
 \caption{\textbf{Estimates of bias and sensitivity for the elephant-rabbit condition.}
\textbf{Left:} Bias, computed as $\mathrm{PSE}-0.5$, reflects shifts in the semantic meaning boundary. The dashed horizontal line indicates the midpoint of the synthesis continuum (zero bias); positive values indicate elephant bias and negative values indicate rabbit bias. 
\textbf{Right:} Higher sensitivity values indicate stronger semantic separation as the mixing parameter $\alpha$ varies.}
    \label{fig:bias_sensitivity_guidance_er}
\end{figure*}

\begin{figure*}[t]
  \centering
  \includegraphics[width=0.9\linewidth]{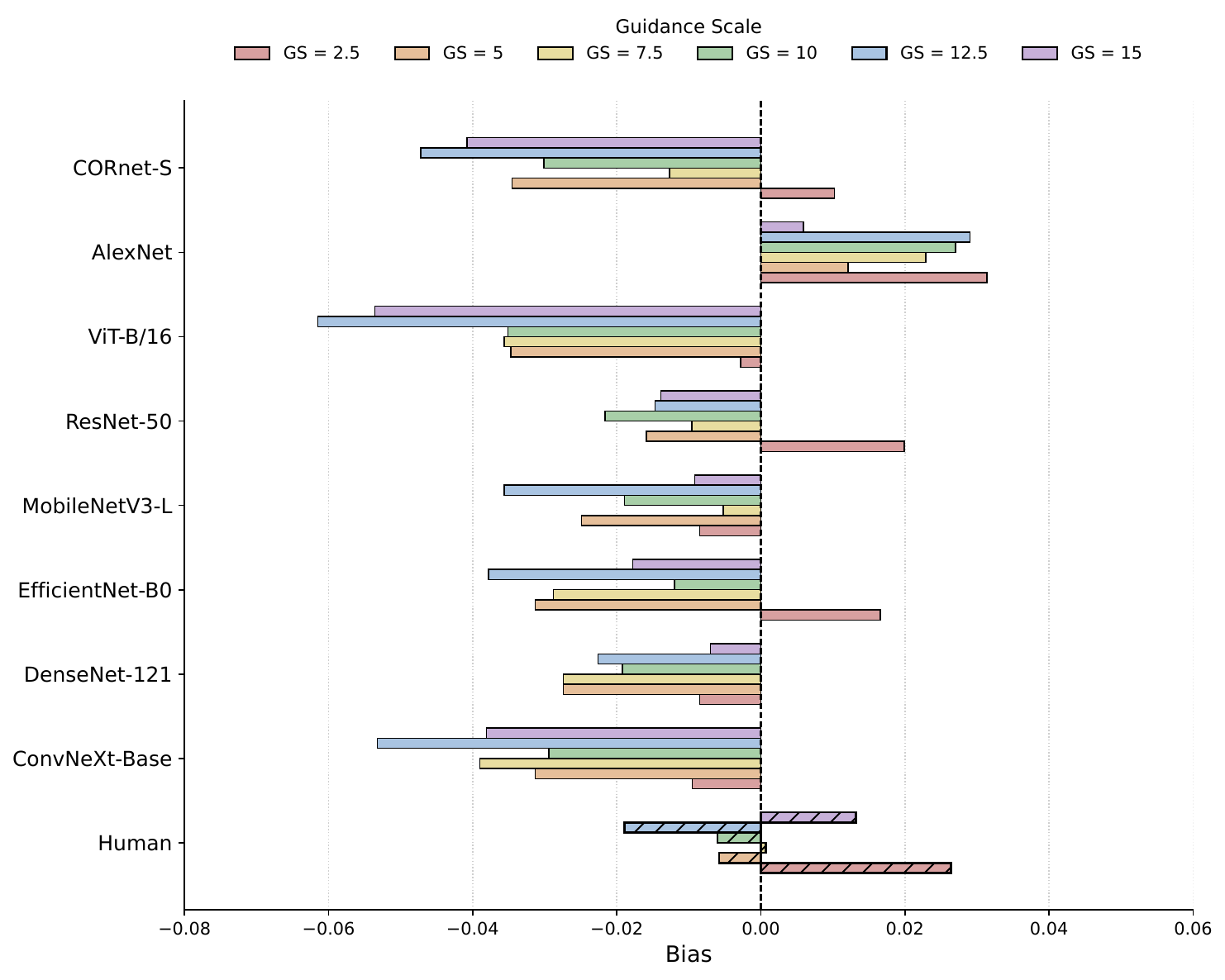} 
  \vspace{-0.6em}
 \caption{\textbf{Bias for the elephant-rabbit condition across guidance scales for humans and classifier models.}
Grouped bars show bias values computed as $\mathrm{PSE}-0.5$ for each model, with colors indicating different guidance scales (GS). 
}\vspace{-0.3em}
    \label{fig:bias_gs_bar_er}
\end{figure*}

\begin{figure*}[t]
  \centering
  \includegraphics[width=\linewidth]{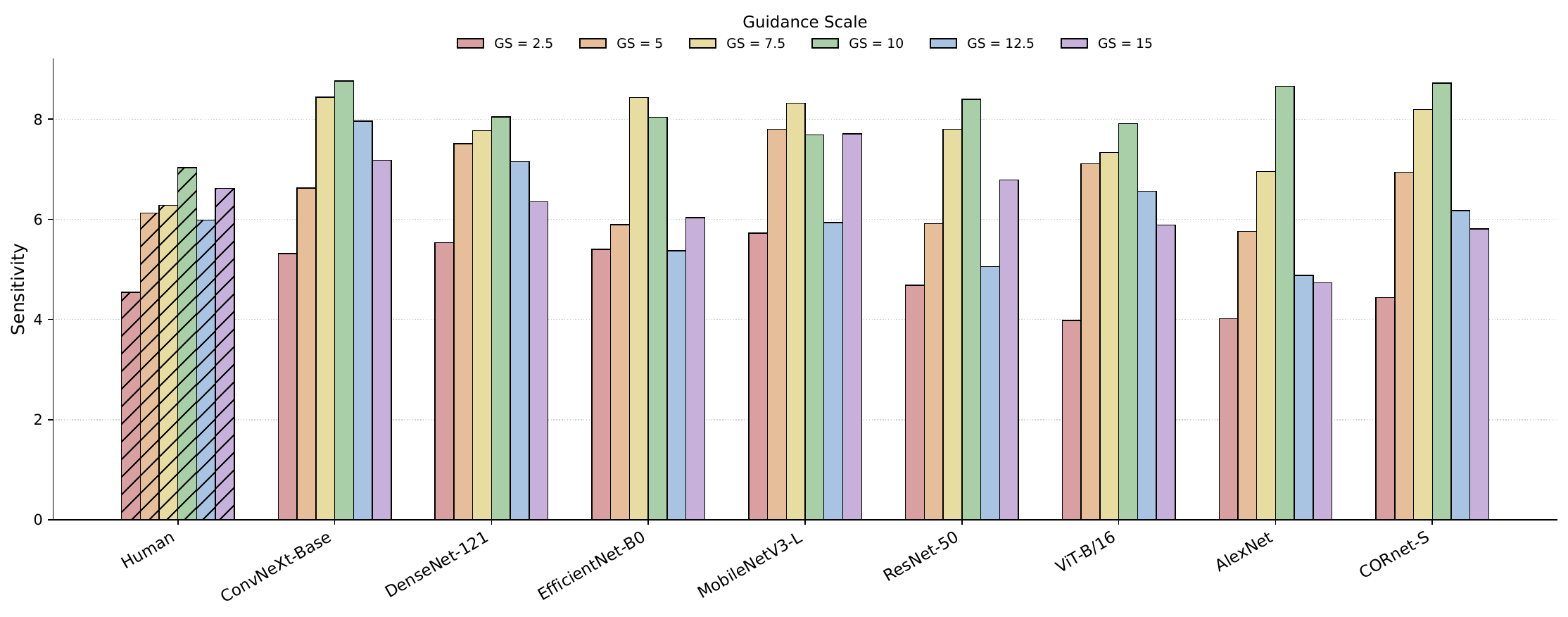}
  \vspace{-0.9em}
\caption{\textbf{Sensitivity for the elephant-rabbit condition across guidance scales for humans and classifier models.}
Higher sensitivity values indicate stronger semantic separation as the mixing parameter $\alpha$ varies.
Grouped bars show sensitivity values for each model, with colors indicating guidance scales (GS).
Higher sensitivity values indicate stronger semantic separation as the mixing parameter $\alpha$ varies.
}
\label{fig:sensitivity_gs_bar_er}
\end{figure*}

In formats analogous to the duck-rabbit condition, we show the 300 elephant-rabbit stimuli (Fig.~\ref{fig:stimuli_generation_elephant_rabbit}), line plots of both bias and sensitivity trends as a function of guidance scale (Fig.~\ref{fig:bias_sensitivity_guidance_er}), tabular summaries of bias (Table~\ref{tab:bias_er}) and sensitivity (Table~\ref{tab:slope_er}), and bar charts of bias (Fig.~\ref{fig:bias_gs_bar_er}) and sensitivity (Fig.~\ref{fig:sensitivity_gs_bar_er}).

Qualitatively, we observe similar patterns to the duck-rabbit condition: humans exhibit lower bias than classifiers, and classifiers exhibit higher sensitivity. However, overall classifier bias is reduced in the elephant-rabbit condition, and the effect of guidance scale is less pronounced. This suggests that the magnitude of classifier bias depends on the specific label pairing, potentially reflecting differences in class priors, feature separability, or asymmetries in the learned representation space. The different machine classifier behavior further suggests that some label pairs may be less sensitive to guidance scale, possibly due to more distinct visual features. Overall, the elephant-rabbit results support the generality of our framework while highlighting that perceptual bias and sensitivity can depend on the specific semantic ambiguity being probed. 

\clearpage
\newpage
\section{Semantic Continuum Endpoint Control Analysis}\label{sec:supp_endpoint}

\begin{figure*}[h]
  \centering
  \includegraphics[width=0.85\linewidth]{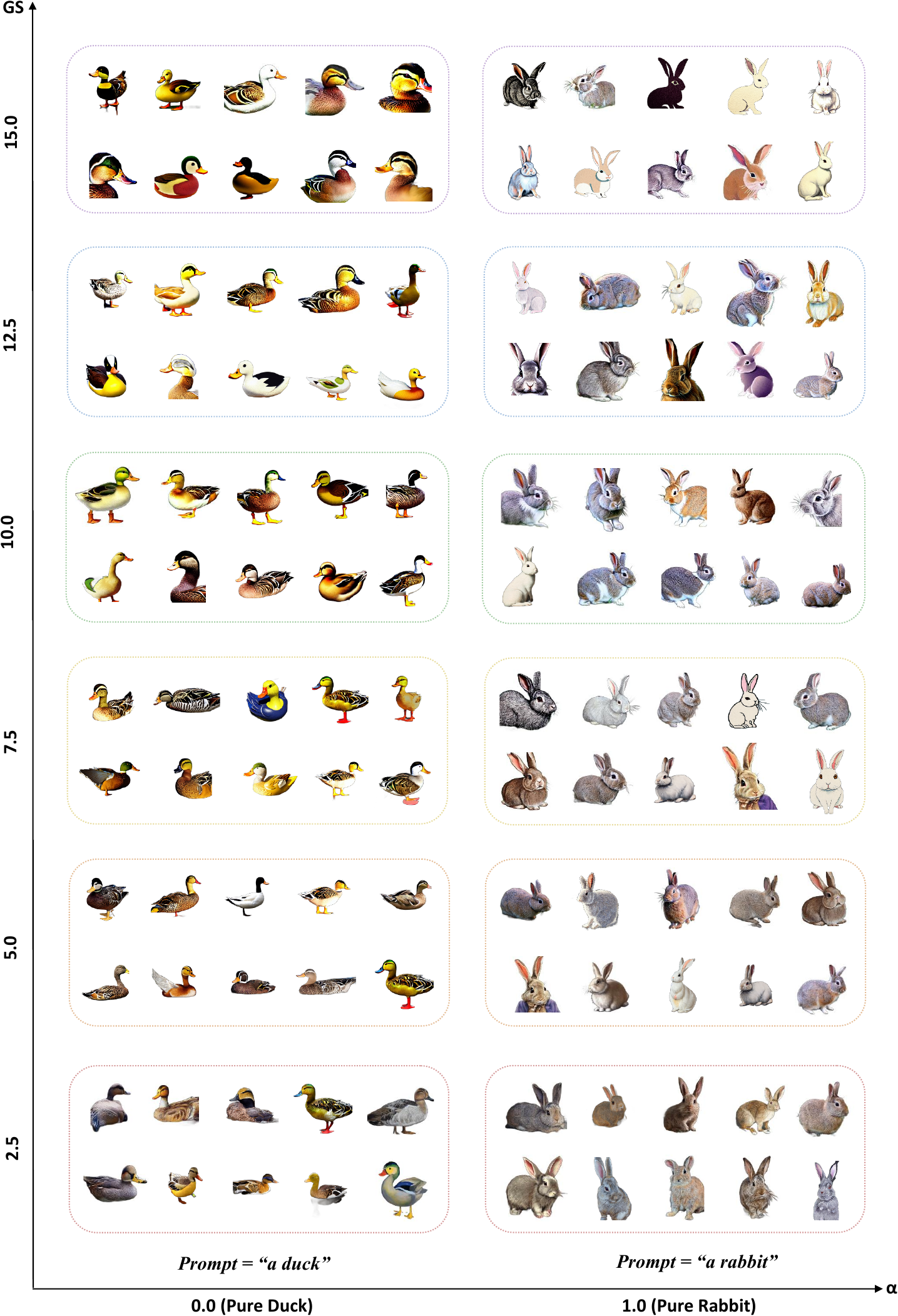}
  \caption{The 120 duck-rabbit stimuli for endpoint control with $\alpha$ = 0 (pure ``duck'') and  $\alpha$  = 1 (pure ``rabbit''). }
  \label{fig:endpoint_dr_stimuli}
\end{figure*}

To assess whether the observed classifier boundary shifts could reflect a general difficulty in distinguishing generated ducks and rabbits, we chose the duck-rabbit condition as an example and evaluated classifier performance at the endpoints of the stimulus continuum: $\alpha$ = 0 (pure ``duck'' prompt) and  $\alpha$  = 1 (pure ``rabbit'' prompt). Endpoint stimuli were generated using the same Stable Diffusion 1.5 pipeline and background processing as the main experiment. We generated 10 images per endpoint at each guidance scale, GS $\in$ $\{2.5,\,5.0,\,7.5,\,10.0,\,12.5,\,15.0\}$, yielding 120 images, shown in Fig.~\ref{fig:endpoint_dr_stimuli}. All eight classifiers were evaluated using their respective preprocessing procedures from the main experiment.

We retained the category aggregation procedure used in the main analysis: ImageNet softmax probabilities were averaged within the selected duck and rabbit category groups and then normalized across the two groups. For this control, images were classified as rabbit when the resulting rabbit-choice probability exceeded 0.5 and as duck otherwise, without Bernoulli sampling. We report endpoint classification accuracy and the mean probability assigned to the correct category, calculated as one minus the rabbit-choice probability for duck endpoints and as the rabbit-choice probability for rabbit endpoints. These probabilities characterize the two-alternative duck–rabbit readout and should not be interpreted as confidence over all 1,000 ImageNet classes.

Seven classifiers correctly classified all 120 endpoint images (100\% accuracy), while AlexNet correctly classified 119 of 120 images (99.17\%). The only misclassification occurred for AlexNet on a rabbit endpoint at GS = 7.5. Across models, the mean correct-category probability over all endpoint images ranged from 97.92\% to 99.94\%. At the lowest guidance scale, GS = 2.5, all eight classifiers correctly classified all 20 endpoint images, with mean correct-category probabilities ranging from 98.95\% to above 99.99\%. Results for each model and guidance scale are reported in the accompanying Table~\ref{endpoint_dr}.

\begin{table}[htbp]
\centering
\resizebox{0.8\linewidth}{!}{
\begin{tabular}{lrrrrrr}
\toprule
\textbf{Model} & \textbf{GS=2.5} & \textbf{GS=5.0} & \textbf{GS=7.5} & \textbf{GS=10.0} & \textbf{GS=12.5} & \textbf{GS=15.0} \\
\midrule
ConvNeXt-Base      & 99.76  & 99.80 & 99.80 & 99.85  & 99.82 & 99.84 \\
ResNet-50          & 99.99  & 99.99 & 99.36 & 100.00 & 99.49 & 99.85 \\
DenseNet-121       & 100.00 & 99.95 & 99.82 & 100.00 & 99.97 & 99.93 \\
MobileNet-V3-Large & 100.00 & 99.93 & 98.34 & 100.00 & 99.52 & 99.36 \\
EfficientNet-B0    & 99.93  & 99.91 & 99.67 & 99.92  & 99.57 & 99.33 \\
ViT-B/16          & 99.94  & 99.97 & 98.47 & 99.97  & 99.87 & 99.78 \\
AlexNet           & 98.95  & 98.85 & 91.47 & 99.99  & 99.37 & 98.89 \\
CORnet-S          & 99.94  & 99.95 & 99.66 & 99.97  & 99.75 & 99.76 \\
\bottomrule
\end{tabular}
}
\vspace{0.6em}
\caption{Mean probability assigned to the correct endpoint category (\%) by model and guidance scale (GS). 
}
\label{endpoint_dr}
\end{table}

These results demonstrate reliable duck–rabbit discrimination on the sampled endpoint images across all tested guidance scales, including the lowest-guidance condition. They therefore argue against a general inability to distinguish generated ducks and rabbits as the sole explanation for the observed classifier boundary shifts.

\end{document}